\documentclass[sn-mathphys,Numbered, iicol]{sn-jnl}%

\usepackage{graphicx}%
\usepackage{multirow}%
\usepackage{amsmath,amssymb,amsfonts}%
\usepackage{amsthm}%
\usepackage{mathrsfs}%
\usepackage[title]{appendix}%
\usepackage{textcomp}%
\usepackage{manyfoot}%
\usepackage{booktabs}%
\usepackage{algorithm}%
\usepackage{algorithmicx}%
\usepackage{algpseudocode}%
\usepackage{listings}%
\usepackage{lipsum}%
\usepackage{titletoc}
\usepackage{bbm}
\usepackage{multirow}
\usepackage{tabularx}

\usepackage[dvipsnames]{xcolor}

\renewcommand{\figurename}{Figure}

\newcommand{\rone}[1]{\textcolor{black}{#1}}
\newcommand{\rtwo}[1]{\textcolor{black}{#1}}
\newcommand{\updated}[1]{\textcolor{black}{#1}}

\DeclareMathSymbol{\shortminus}{\mathbin}{AMSa}{"39}

\definecolor{flodarkpurple}{rgb}{0.288,0.1196,0.7}
\newcommand{\graytext}[1]{\textcolor{flodarkpurple}{#1}}

\usepackage{inconsolata}
\usepackage[T1]{fontenc}

\definecolor{light-gray}{rgb}{0.8, 0.8, 0.8}
\definecolor{comment-green}{rgb}{0.435, 0.576, 0.106}
\definecolor{prompt-blue}{HTML}{2596be}
\definecolor{code-function}{HTML}{379fbe}
\definecolor{code-function}{HTML}{693da8}  %
\definecolor{code-syntax}{HTML}{0060b1}
\definecolor{code-constant}{HTML}{d86001}
\definecolor{prompt-gray}{HTML}{a7a7a7}
\definecolor{highlight}{HTML}{f8f9cb}
\definecolor{highlight}{HTML}{e3eeff}  %
\definecolor{code-perception}{HTML}{2ecc71}
\definecolor{code-control}{HTML}{ff9900}
\definecolor{code-undefined}{HTML}{ff0000}
\renewcommand\fbox{\fcolorbox{light-gray}{white}}

\newcommand{\link}[2]{\textcolor{magenta}{\href{#1}{#2}}}

\NewDocumentCommand{\code}{v}{%
\texttt{\small{\textcolor{code-syntax}{#1}}}%
}

\newcommand{\gs}{\textbf{\textsc{generator-scorer}}}
\newcommand{\scgs}{\textbf{\textsc{saycan-gs}}}
\newcommand{\imgs}{\textbf{\textsc{innermono-gs}}}
\newcommand{\sh}{\textbf{\textsc{shooting}}}
\newcommand{\se}{\textbf{\textsc{greedy-search}}}
\newcommand{\ttm}{\textbf{Text2Motion}}

\newcommand{\scgsnb}{\textsc{saycan-gs}}
\newcommand{\imgsnb}{\textsc{innermono-gs}}
\newcommand{\shnb}{\textsc{shooting}}
\newcommand{\senb}{\textsc{greedy-search}}
\newcommand{\ttmnb}{Text2Motion}
\newcommand{\LH}{\textbf{LH}}
\newcommand{\LG}{\textbf{LG}}
\newcommand{\PAP}{\textbf{PAP}}

\newcommand{\func}[2]{\mathop{}#1\left(#2\right)}
\newcommand{\E}[2]{\mathop{}\operatorname{E}_{#1}\left[#2\right]}
\newcommand{\V}[2]{\mathop{}\operatorname{Var}_{#1}\left[#2\right]}
\DeclareMathOperator*{\argmax}{arg\,max}

\begin{document}

\title[Article Title]{\textbf{Text2Motion}: From Natural Language Instructions to Feasible Plans}

\author*[1]{\fnm{Kevin} \sur{Lin}}\email{kevin.lin@cs.stanford.edu}

\author*[1,2]{\fnm{Christopher} \sur{Agia}}\email{cagia@stanford.edu}

\author[1]{\fnm{Toki} \sur{Migimatsu}}\email{takatoki@cs.stanford.edu}

\author[2]{\fnm{Marco} \sur{Pavone}}\email{pavone@stanford.edu}

\author[1]{\fnm{Jeannette} \sur{Bohg}}\email{bohg@stanford.edu}

\affil[1]{\orgdiv{Department of Computer Science}, \orgname{Stanford University}, \orgaddress{\state{California}, \country{U.S.A}}}

\affil[2]{\orgdiv{Department of Aeronautics \& Astronautics}, \orgname{Stanford University}, \orgaddress{\state{California}, \country{U.S.A}}}

\abstract{
We propose Text2Motion, a language-based planning framework enabling robots to solve sequential manipulation tasks that require long-horizon reasoning. 
Given a natural language instruction, our framework constructs both a task- and motion-level plan that is verified to reach inferred symbolic goals.
Text2Motion uses feasibility heuristics encoded in Q-functions of a library of skills to guide task planning with Large Language Models.
Whereas previous language-based planners only consider the feasibility of individual skills, Text2Motion actively resolves geometric dependencies spanning skill sequences by performing geometric feasibility planning during its search.
We evaluate our method on a suite of problems that require long-horizon reasoning, interpretation of abstract goals, and handling of partial affordance perception. 
Our experiments show that Text2Motion can solve these challenging problems with a success rate of 82\%, while prior state-of-the-art language-based planning methods only achieve 13\%.
Text2Motion thus provides promising generalization characteristics to semantically diverse sequential manipulation tasks with geometric dependencies between skills.
Qualitative results are made available at \link{https://sites.google.com/stanford.edu/text2motion}{sites.google.com/stanford.edu/text2motion}.
}


\maketitle

\section{Introduction}\label{sec:introduction}
Long-horizon robot planning is traditionally formulated as a joint symbolic and geometric reasoning problem, where the symbolic reasoner is supported by a formal logic representation (e.g. first-order logic~\cite{aeronautiques1998pddl}).
Such systems can generalize within the logical planning domain specified by experts. 
However, many desirable properties of plans that can be \rone{conveniently expressed in language by non-expert users} may be cumbersome to specify in formal logic.
Examples include the specification of user intent or preferences.

The emergence of {\em Large Language Models} (LLMs)~\cite{foundation-models-2021} as a task-agnostic reasoning module presents a promising pathway to general robot planning capabilities.
Several recent works~\cite{saycan-2022, innermono-2022, code-as-policies-2022, wu2023tidybot} capitalize on their ability to perform task planning for robot systems without needing to manually specify symbolic planning domains.
\rone{
Nevertheless, these prior approaches adopt myopic or open-loop execution strategies, trusting LLMs to produce correct plans without verifying them on the symbolic or geometric level.
Such strategies are challenged in long-horizon settings, where the task planning abilities of even the most advanced LLMs appear to degrade~\cite{llms-cant-plan-2022}, and the overall success of a seemingly correct task plan depends as well on how it is executed to ensure long-horizon feasibility.
Therefore, we ask in this paper: how can we verify the correctness and feasibility of LLM-generated plans prior to execution?}

We propose \ttm{}, a language-based planning framework that interfaces an LLM with a library of learned skills and a geometric feasibility planner~\cite{taps-2022} to solve complex sequential manipulation tasks (Figure~\ref{fig:teaser}).
Our contributions are two-fold: (i) a hybrid LLM planner that synergistically integrates shooting-based and search-based planning strategies to construct geometrically feasible plans for tasks not seen by the skills during training; and (ii) a plan termination method that infers goal states from a natural language instruction to verify the completion of plans before executing them. 
We find that our planner achieves a success rate of 82\% on a suite of challenging table top manipulation tasks, while prior language-based planning methods achieve a 13\% success rate.

\section{Related Work}\label{sec:related-work}
\subsection{Language for robot planning}
\label{subsec:language-literature}
Language is increasingly being explored as a medium for solving long-horizon robotics problems.
For instance, {\em Language-conditioned policies\/} (LCPs) are not only used to learn short-horizon skills~\cite{stepputtis2020language, jang2021bczero, concept2robot-2021, cliport-2022, perceiver-2022, vima-2022}, but also long-horizon policies~\cite{mees2022calvin, rt1-2022, dalal2023imitating}.
However, LCPs require expensive data collection and training procedures if they are to generalize to a wide distribution of long-horizon tasks with diverse instructions.

Several recent works leverage the generative qualities of LLMs by prompting them to predict long-horizon plans.
\cite{zeroshot-llms-2022} grounds an LLM planner to admissible action sets for task planning, \cite{silver2022pddl, liu2023llm+} explore the integration of LLMs with PDDL~\cite{aeronautiques1998pddl}, and \cite{wang2023describe, skreta2023errors} focuses on task-level replanning with LLMs.
Tangential works shift the representation of plans from action sequences to code~\cite{code-as-policies-2022, progprompt-2022, zelikman2022parsel, vemprala2023chatgpt} and embed task queries, robot actions, solution samples, and fallback behaviors as programs in the prompt. 
In contrast to these works, which primarily address challenges in task planning with LLMs, we focus on verifying LLM-generated plans for feasibility on the geometric level.

\begin{figure}
    \centering
    \includegraphics[width=0.48\textwidth]{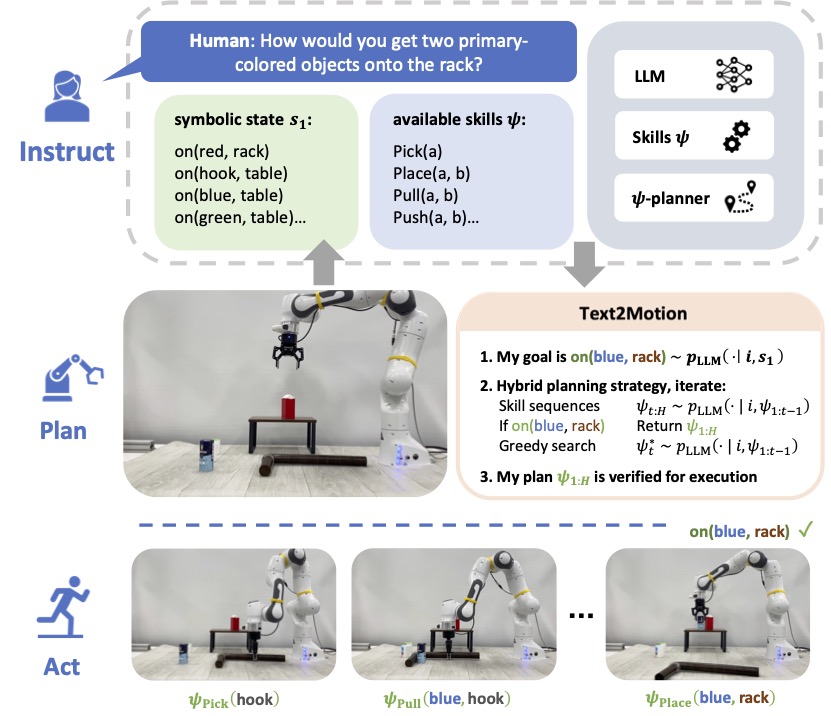}
    \vspace{-5pt}
    \caption{To carry out the instruction ``get two primary-colored objects onto the rack,'' the robot must apply symbolic reasoning over the scene description and language instruction to deduce what skills should be executed to acquire a second primary-colored object, after noticing that a red object is already on the rack (i.e. \graytext{on(red, rack)}).
    It must also apply geometric reasoning to ensure that skills are sequenced in a manner that is likely to succeed. 
    Unlike prior work~\cite{saycan-2022, innermono-2022} that myopically executes skills at the current timestep, \ttmnb{} constructs \textit{sequences} of skills and coordinates their geometric dependencies with geometric feasibility planning~\cite{taps-2022}. 
    Upon planning the skill sequence \graytext{Pick(hook)}, \graytext{Pull(blue, hook)}, \graytext{Pick(blue)}, \graytext{Place(blue, rack)}, our method computes a grasp position on the hook that enables pulling the blue object into the robot workspace so that it can be successfully picked up in the next step.}
    \label{fig:teaser}
\end{figure}

Closest in spirit to our work are SayCan~\cite{saycan-2022} and Inner Monologue (IM)~\cite{innermono-2022} which at each timestep score the \textit{usefulness} and \textit{feasibility} of all possible skills and execute the one with the highest score.
Termination occurs when the score of the $\texttt{stop}$ ``skill'' is larger than any other.
IM provides additional sources of feedback to the LLM in the form of skill successes and task-progress cues. 

While SayCan and IM are evaluated on a diverse range of tasks, there are several limitations that impede their performance in the settings we study.
First, by only myopically executing the next skill at each timestep, they may fail to account for geometric dependencies that exist over the extent of a skill sequence. 
For an example, see Figure~\ref{fig:teaser}.
Second, they do not explicitly predict a multi-step plan, which prevents verification of desired properties or outcomes prior to execution.
Examples of such properties could include whether the final state induced by the plan satisfies symbolic constraints or whether the plan adheres to safety criteria.
Lastly, these methods ignore the uncertainty of skill feasibility predictions (i.e. affordances), which \cite{taps-2022} demonstrates is important when sequencing learned skills to solve long-horizon tasks.
By addressing these limitations, \ttm{} outperforms SayCan and IM by a large margin on tasks with geometric dependencies, as demonstrated in the experiments.

\subsection{Task and motion planning}
\label{subsec:tamp-literature}
Task and Motion Planning (TAMP) refers to a problem setting in which a robot solves long-horizon tasks through symbolic and geometric reasoning~\cite{kaelbling2012integrated, integrated-tamp-2021}.
The hierarchical approach~\cite{5980391} characterizes the most common family of solution methods.
Such methods typically employ a) a symbolic task planner~\cite{bonet2001planning, helmert2006fast} to produce a candidate plan skeleton, and b) a motion planner to verify the plan skeleton for its geometric feasibility and compute a motion trajectory subject to robot and environmental constraints~\cite{garrett2020-pddlstream, toussaint2015-lgp, driess2019-hlgp}.

For complex tasks, classical TAMP solvers~\cite{kaelbling2011-HPN, lagriffoul2014efficiently, toussaint2015-lgp, dantam2016incremental, bidot2017geometric, garrett2020-pddlstream} may iterate between task planning and motion planning for minutes until a plan is found.
To amortize planning costs, works learn sampling distributions~\cite{wang2018active, xu2021deep, kim2019adversarial, kim2022representation, taps-2022}, visual feasibility heuristics~\cite{driess2020-dvr, driess2020-dvh, driess2021-lpr}, low-level controllers~\cite{driess2021-lgr, silver2022learning}, or state sparsifiers~\cite{chitnis2021camps, silver2021-ploi}, from datasets of solutions computed by classical TAMP solvers.
Another line of works learn symbolic representations for TAMP~\cite{kroemer2016learning, ames2018learning, konidaris2018skills, silver2021learning, wang2021learning, curtis2022discovering, chitnis2022learning, silver2022learning}, often from task-specific symbolic transition experience.

\rone{
As is common in TAMP, \ttm{} also assumes knowledge of task-relevant objects and their poses in order to plan feasible trajectories for long-horizon tasks.
However, central to our work is the use of LLMs instead of symbolic task planners often used in TAMP~\cite{integrated-tamp-2021}, and language as convenient medium to express tasks that may be cumbersome to specify in formal logic (e.g. user preferences~\cite{wu2023tidybot}). 
Accordingly, we address early challenges concerning the reliable use of LLMs (discussed in Section~\ref{subsec:language-literature}) in the long-horizon settings typically solved by TAMP.
\ttm{} thereby presents several qualitative differences from TAMP: i) the ability to interpret free-form language instructions for the construction of multi-step plans, and ii) the capacity to reason over an unrestricted set of object classes and object properties, both of which are supported by the commonsense knowledge of LLMs~\cite{brown2020language}.
We leave the extension of our framework to open-world settings (e.g. via environment exploration~\cite{ chen2022open} or interaction~\cite{curtis2022-unknown}) to future work.}

\section{Problem Setup}
\label{sec:problem-setup}
We aim to solve long-horizon sequential manipulation problems that require symbolic and geometric reasoning from a natural language instruction $i$ and the initial state of the environment $s_1$. 
\rtwo{We assume a closed-world setting, whereby the initial state $s_1$ contains knowledge of task-relevant objects and their poses as provided by an external perception system (Appendix~\ref{appx-sub:hardware-setup}).
Fulfillment of the instruction $i$ corresponds to achieving a desired goal configuration of the task-relevant objects which can be symbolically expressed with a closed set of predicates (Appendix~\ref{appx-sub:scene-descr-symbolic}).}

\subsection{LLM and skill library}
\label{subsec:prelminaries}
We assume access to an LLM and a library of skills $\mathcal{L}^\psi = \{\psi^1, \ldots, \psi^N\}$.
Each skill $\psi$ consists of a policy $\pi(a | s)$ and a parameterized manipulation primitive $\phi(a)$~\cite{felip2013manipulation}, and is associated with a contextual bandit, or a single-timestep Markov Decision Process (MDP):
\begin{equation}
    \label{eq:skill-mdp}
    \mathcal{M} = (\mathcal{S}, \mathcal{A}, T, R, \rho), 
\end{equation}
where $\mathcal{S}$ is the state space, $\mathcal{A}$ is the action space, $T(s' | s, a)$ is the transition model, $R(s, a, s')$ is the binary reward function, and $\rho(s)$ is the initial state distribution.
When a skill $\psi$ is executed, an action $a \in \mathcal{A}$ is sampled from its policy $\pi(a|s)$ and fed to its primitive $\phi(a)$, which consumes the action and executes a series of motor commands on the robot. 
If the skill succeeds, it receives a binary reward of $r$ (or $\neg r$ if it fails).
We subsequently refer to policy actions $a \in \mathcal{A}$ as \textit{parameters} for the primitive, which,  depending on the skill, can represent grasp poses, placement locations, and pulling or pushing distances (Appendix~\ref{appx-sub:learning-models}).

A timestep in our environment corresponds to the execution of a single skill.
We assume that each skill comes with a language description and that methods exist to obtain its policy $\pi(a|s)$, Q-function $Q^\pi(s, a)$, and dynamics model $T^\pi(s' | s, a)$. 
Our framework is agnostic to the approach used to obtain these models.
We also assume a method to convey the environment state $s \in \mathcal{S}$ to the LLM as natural language. 

\subsection{The planning objective}
\label{subsec:planning-objective}
Our objective is to find a plan in the form of a sequence of skills $[\psi_1, \ldots, \psi_H]$ (for
notational convenience, we hereafter represent sequences
with range subscripts, e.g. $\psi_{1:H}$) that is both likely to satisfy the instruction $i$ and can be successfully executed from the environment's initial state $s_1$. 
This objective can be expressed as the joint probability of skill sequence $\psi_{1:H}$ and binary rewards $r_{1:H}$ given the instruction $i$ and initial state $s_1$:
\begin{equation}
    \begin{split}
        &p(\psi_{1:H}, r_{1:H} \mid i, s_1) \\
        &\quad\quad= p(\psi_{1:H} \mid i, s_1)\, p(r_{1:H} \mid i, s_1, \psi_{1:H}).
    \end{split} \label{eq:tamp-score}
\end{equation}
The first term in this product $p(\psi_{1:H} \mid i, s_1)$ considers the probability that the skill sequence $\psi_{1:H}$ will satisfy the instruction $i$ from a symbolic perspective. 
However, a symbolically correct skill sequence may fail during execution due to kinematic constraints of the robot or geometric dependencies spanning the skill sequence.
We must also consider the \textit{success probability} of the skill sequence $\psi_{1:H}$ captured by the second term in this product $p(r_{1:H} \mid i, s_1, \psi_{1:H})$.
The success probability depends on the parameters $a_{1:H}$ fed to the underlying sequence of primitives $\phi_{1:H}$ that control the robot's motion:
\begin{equation}
    \label{eq:motion-parameter-score}
    p(r_{1:H} \mid i, s_1, \psi_{1:H}) = p(r_{1:H} \mid s_1, a_{1:H}).
\end{equation}
Eq.~\ref{eq:motion-parameter-score} represents the probability that skills $\psi_{1:H}$ achieve rewards $r_{1:H}$ when executed from initial state $s_1$ with parameters $a_{1:H}$; which is independent of the instruction $i$.
If just one skill fails (reward $\neg r$), then the entire plan fails.

\subsection{Geometric feasibility planning}
\label{subsec:geometric-feasibility-planning}
The role of geometric feasibility planning is to maximize the success probability (Eq.~\ref{eq:motion-parameter-score}) of a skill sequence $\psi_{1:H}$ by computing an optimal set of parameters $a_{1:H}$ for the underlying primitive sequence $\phi_{1:H}$.
This process is essential for finding plans that maximize the overall planning objective in Eq.~\ref{eq:tamp-score}.
In our experiments, we leverage Sequencing Task-Agnostic Policies (STAP)~\cite{taps-2022}.

STAP resolves geometric dependencies across the skill sequence $\psi_{1:H}$ by maximizing the product of step reward probabilities of parameters $a_{1:H}$:
\begin{equation}
    \label{eq:taps-objective}
    a_{1:H}^* = \arg \max_{a_{1:H}} \, \E{s_{2:H}}{\prod_{t=1}^H p(r_t \mid s_t, a_t)},
\end{equation}
where future states $s_{2:H}$ are predicted by dynamics models $s_{t+1} \sim T^{\pi_t}(\cdot | s_t, a_t)$. 
Note that the reward probability $p(r_t \mid s_t, a_t)$ is equivalent to the Q-function $Q^{\pi_t}(s_t, a_t)$ for skill $\psi_t$ in a contextual bandit setting with binary rewards (Eq.~\ref{eq:skill-mdp}).
The success probability of the optimized skill sequence $\psi_{1:H}$ is thereby approximated by the product of Q-functions evaluated from initial state $s_1$ along a sampled trajectory $s_{2:H}$ with parameters $a^*_{1:H}$:
\begin{equation}
    \label{eq:stap-score}
    p(r_{1:H} \mid s_1, a_{1:H}) \approx \prod_{t=1}^H Q^{\pi_t}(s_t, a^*_t).
\end{equation}

In principle, our framework is agnostic to the specific approach used for geometric feasibility planning, requiring only that it is compatible with the skill formalism defined in Section~\ref{subsec:prelminaries} and provides a reliable estimate of Eq.~\ref{eq:motion-parameter-score}.

\section{Methods}
\label{sec:text2motion}

The core idea of this paper is to ensure the geometric feasibility of an LLM task plan---and thereby its correctness---by predicting the success probability (Eq.~\ref{eq:motion-parameter-score}) of learned skills that are sequenced according to the task plan.
In the following sections, we outline two strategies for planning with LLMs and learned skills: a shooting-based planner and a search-based planner.
We then introduce the full planning algorithm, \ttm{}, which synergistically integrates the strengths of both strategies.
These strategies represent different ways of maximizing the overall planning objective in Eq.~\ref{eq:tamp-score}.

\begin{figure*}
    \centering
     \vspace{-30pt}
    \includegraphics[width=0.98\textwidth]{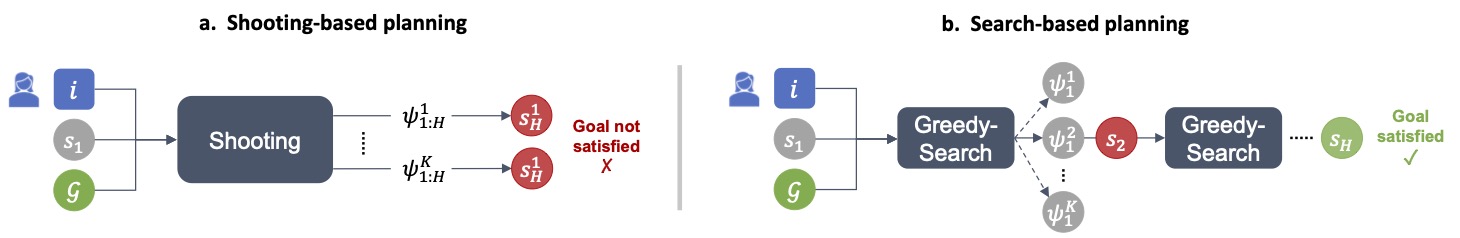}
    \caption{\textbf{\shnb{} and \senb{} planning overview}. 
    Both \shnb{} and \senb{} planners use the LLM to predict the set of valid goal states given the user's natural language instruction and a description of the current state of the environment. \updated{These predicted goals are used to decide when the instruction is satisfied and planning can terminate. \textbf{Left:} The \shnb{} strategy uses the LLM to propose full skill sequences first and then runs geometric feasibility planning afterwards. As shown in the experiments, this approach fails when the space of candidate task plans is large but few skill sequences are geometrically feasible. \textbf{Right:} In the \senb{} strategy, the LLM is used to propose $K$ candidate skills with the top LLM scores. The geometric feasibility planner then evaluates the feasibility of each candidate skill, and the one with the highest product of LLM and geometric feasibility scores is selected. The successor state of this skill is predicted by the geometric feasibility planner's dynamics model. If the successor state does not satisfy any of the predicted goals, then it is given to the LLM to plan the next skill. If a goal is satisfied, then the planner returns the skill sequence for execution. By interleaving LLM task planning with geometric feasibility planning at each planning iteration, \senb{} is able to reliably find feasible plans across the different families of tasks we study in the experiments.}
    }
    \label{fig:system}
\end{figure*}

\subsection{Goal prediction} 
\label{subsec:goal-prediction}
Plans with high overall objective scores (Eq.~\ref{eq:tamp-score}) are not guaranteed to satisfy their instruction.
Consider the instruction ``move all the dishes from the table to the sink'' issued in an environment with two dishes on the table. 
While a plan that picks and places one of the two dishes in the sink may have a high language model likelihood and success probability, it fails to satisfy the instruction. 

The first step in all planning strategies is to convert the language instruction into a goal condition that can be checked against a candidate sequence of skills.
\rone{Given an instruction $i$, a set of objects $O$ in the scene, and a library of predicate classifiers $\mathcal{L}^\chi=\{\chi^1,\ldots,\chi^M\}$, we use the LLM to predict a set of $|\mathcal{G}|$ symbolic goal propositions $\mathcal{G} = \{g_1, \ldots, g_j\}$ that would satisfy the instruction. 
Each goal proposition $g\in\mathcal{G}$ is a set of predicates grounded over objects in the scene.
Each predicate is a binary-valued function over objects and has a one-to-one correspondence with a predicate classifier $\chi\in\mathcal{L}^\chi$ that implements the predicate (details in Appendix~\ref{appx-sub:scene-descr-symbolic}).
We define a satisfaction function $\func{F^\mathcal{G}_{\text{sat}}}{s}: \mathcal{S} \rightarrow \{0, 1\}$ which takes as input a geometric state $s$ and evaluates to $1$ if any goal proposition $g\in\mathcal{G}$ predicted by the LLM holds in state $s$.}

A sequence of skills $\psi_{1:H}$ is said to satisfy the instruction $i$ \textit{iff}:
\begin{equation}
    \label{eq:satisfaction-function}
    \exists\: s \in s_{2:H+1}: F^\mathcal{G}_{\text{sat}}(s) = 1,
\end{equation}
where the future states $s_{2:H+1}$ are predicted by the geometric feasibility planner (see Section~\ref{subsec:geometric-feasibility-planning}).
If $F^\mathcal{G}_{\text{sat}}(s_t)$ evaluates to $1$ for a geometric state $s_t$ at timestep $t \le H+1$, then the planner returns the subsequence of skills $\psi_{1:t-1}$ for execution.

\begin{algorithm}
\updated{
\small
\caption{Shooting-based LLM planner}\label{alg:shooting}
\begin{algorithmic}[1]
\State \textbf{globals:} $\mathcal{L}^\psi, \mathcal{L}^\chi, \textsc{SatFunc}, \textsc{LLM}, \textsc{STAP}$
\Function{Shooting}{$i, s_1, \mathcal{G}; K$}
    \State $F^\mathcal{G}_{\text{sat}} \gets \Call{SatFunc}{\mathcal{G}, \mathcal{L}^\chi}$ \Comment{Goal checker}
    \State $\{\psi^{(j)}_{1:H}\}_{j=1}^K \gets \Call{LLM}{i, s_1, \mathcal{G}, K}$ \Comment{Gen. plans}
    \State $C = \{\,\}$ \Comment{Init. candidate set}
    \For{$j = 1 \ldots K$}
        \State $s^{(j)}_{2:H+1}, a^{(j)}_{1:H} \gets \Call{STAP}{s_1, \psi^{(j)}_{1:H}, \mathcal{L}^\psi}$ 
        \If{$F^\mathcal{G}_{\text{sat}}(s^{(j)}_t) == 1$ \textbf{for} $t \leq H+1$}
            \State $\psi^{(j)}_{1:t-1} \gets \psi^{(j)}_{1:H}[:t-1]$ \Comment{Slice plan}
            \State $C \gets C \cup \{j\}$ \Comment{Add to candidate set}
        \EndIf
        \State Compute $p_{\text{success}}^{(j)}$ via Eq.~\ref{eq:stap-score}
    \EndFor
    \State Filter OOD plans from $C$ as per Eq.~\ref{eq:ood-detection}
    \If{$C == \emptyset$}
        \State \textbf{raise} \texttt{planning failure}
    \EndIf
    \State $j^* = \argmax_{j \in C}\; p_{\text{success}}^{(j)}$
    \State \Return $\psi^{(j^*)}_{1:t-1}$ \Comment{Return best plan}
\EndFunction
\end{algorithmic}
}
\end{algorithm}

\subsection{Shooting-based planning}
\label{sec:shooting}
The planner is responsible for finding geometrically feasible plans that satisfy the goal condition predicted by the LLM (Section~\ref{subsec:goal-prediction}).
To this end, the first strategy we propose is a shooting-based planner, termed \sh{} (see Figure~\ref{fig:system}, Left), which takes a single-step approach to maximizing the overall planning objective in Eq.~\ref{eq:tamp-score}.
\sh{}'s process is further outlined in Algorithm~\ref{alg:shooting}.

\sh{} requires querying the LLM only once to generate $K$ candidate skill sequences $\{\psi^{1}_{1:H}, \ldots, \psi^{K}_{1:H}\}$ in an open-ended fashion.
Each candidate skill sequence is processed by the geometric feasibility planner which returns an estimate of the sequence's success probability (Eq.~\ref{eq:stap-score}) and its predicted future state trajectory ${s_{2:H+1}}$.
Skill sequences that satisfy the goal condition (Eq.~\ref{eq:satisfaction-function}) are added to a candidate set. 
Invalid skill sequences as determined by Section~\ref{subsec:ood-detection} are filtered-out of the candidate set.
If the candidate set is not empty, \sh{} returns the skill sequence with the highest success probability, or raises a \texttt{planning failure} otherwise.

\subsection{Search-based planning}
\label{sec:greedy-search}
We propose a second planner, \se{} (see Figure ~\ref{fig:system}, Right), which at each planning iteration ranks candidate skills predicted by the LLM and adds the top scoring skill to the running plan.

This iterative approach can be described as a decomposition of the planning objective in Eq.~\ref{eq:tamp-score} by timestep $t$:
\begin{equation}
    \label{eq:tamp-score-decomp}
    \begin{split}
        &p(\psi_{1:H}, r_{1:H} \mid i, s_1) \\
        &\quad\quad= \prod_{t=1}^H p(\psi_t, r_t \mid i, s_1, \psi_{1:t-1},  r_{1:t-1}).
    \end{split}
\end{equation}
We define the joint probability of $\psi_t$ and $r_t$ in Eq.~\ref{eq:tamp-score-decomp} as the skill score $S_{\text{skill}}$:
\begin{equation*}
    \label{eq:tamp-step-score}
    S_{\text{skill}}(\psi_t) 
        = p(\psi_t, r_t \mid i, s_1, \psi_{1:t-1}, r_{1:t-1}),
\end{equation*}
which we factor using conditional probabilities:
\begin{equation}
    \label{eq:tamp-step-score-factor}
    \begin{split}
        S_{\text{skill}}(\psi_t) &= p(\psi_t \mid i, s_1, \psi_{1:t-1}, r_{1:t-1}) \\ 
        &\quad\quad\quad\quad p(r_t \mid i, s_1, \psi_{1:t}, r_{1:t-1}).
    \end{split}
\end{equation}
Each planning iteration of \se{} is responsible for finding the skill $\psi_t$ that maximizes the skill score (Eq.~\ref{eq:tamp-step-score-factor}) at timestep $t$. 

\textbf{Skill usefulness:}
The first factor of Eq.~\ref{eq:tamp-step-score-factor} captures the \textit{usefulness} of a skill generated by the LLM with respect to satisfying the instruction. 
We define the skill usefulness score $S_{\text{llm}}$:
\begin{align}
    S_{\text{llm}}(\psi_t) 
        &= p(\psi_t \mid i, s_1, \psi_{1:t-1}, r_{1:t-1}) \label{eq:llm-step-score} \\
        &\approx p(\psi_t \mid i, s_{1:t}, \psi_{1:t-1}). \label{eq:llm-step-score-decomp}
\end{align}
In Eq.~\ref{eq:llm-step-score-decomp}, the probability of the next skill $\psi_t$ (Eq.~\ref{eq:llm-step-score}) is cast in terms of the predicted state trajectory $s_{2:t}$ of the running plan $\psi_{1:t-1}$, and is thus is independent of prior rewards $r_{1:t-1}$. 
We refer to Appendix~\ref{appx-sub:skill-usefulness} for a detailed derivation of Eq.~\ref{eq:llm-step-score-decomp}.

At each planning iteration $t$, we optimize $S_{\text{llm}}(\psi_t)$ by querying an LLM to generate $K$ candidate skills $\{\psi_t^1, \dots, \psi_t^K\}$. 
We then compute the usefulness scores $S_{\text{llm}}(\psi_t^k)$ by summing the token log-probabilities of each skill's language description \rtwo{(visualized in Section~\ref{subsec:prompt-engineering})}.
These scores represent the likelihood that $\psi^k_t$ is the correct skill to execute from a language modeling perspective to satisfy instruction $i$.

\begin{algorithm}[t]
\small
\updated{
\caption{Search-based LLM planner}\label{alg:greedy-search}
\begin{algorithmic}[1]
\State \textbf{globals:} $\mathcal{L}^\psi, \mathcal{L}^\chi, \textsc{SatFunc}, \textsc{LLM}, \textsc{STAP}$
\Function{Greedy-Search}{$i, s_1, \mathcal{G}; K, d_{\text{max}}$}
    \State $F^\mathcal{G}_{\text{sat}} \gets \Call{SatFunc}{\mathcal{G}, \mathcal{L}^\chi}$ \Comment{Goal checker}
    \State $\Psi = [\,]$; $\tau = [s_1]$ \Comment{Init. running plan}
    \While{$len(\Psi) < d_{\text{max}}$}
        \State $\Psi, \tau \gets \Call{Greedy-Step}{i, s_1, \mathcal{G}, \Psi, \tau, K}$
        \If{$F^\mathcal{G}_{\text{sat}}(\tau[-1]) == 1$} 
            \State \Return $\Psi$ \Comment{Return goal-reaching plan}
        \EndIf
    \EndWhile
    \State \textbf{raise} \texttt{planning failure}
\EndFunction
\Function{Greedy-Step}{$i, s_1, \mathcal{G}, \Psi, \tau; K$}
    \State $t = len(\Psi) + 1$ \Comment{Curr. planning iteration}
    \State $\{\psi^{(j)}_t\}_{j=1}^K \gets \Call{LLM}{i, \tau, \mathcal{G}, K}$ \Comment{Gen. skills}
    \State $C = \{\,\}$ \Comment{Init. candidate set}
     \For{$j = 1 \ldots K$}
        \State $\psi^{(j)}_{1:t} \gets \Psi.append(\psi^{(j)})$
        \State $s^{(j)}_{2:t+1}, a^{(j)}_{1:t} \gets \Call{STAP}{s_1, \psi^{(j)}_{1:t}, \mathcal{L}^\psi}$ 
        \State Compute $S_{\text{llm}}(\psi^{(j)}_t)$ via Eq.\ref{eq:llm-step-score-decomp}
        \State Compute $S_{\text{geo}}(\psi^{(j)}_t)$ via Eq.\ref{eq:motion-step-score-decomp}
        \State $S_{\text{skill}}(\psi^{(j)}_t) \gets S_{\text{llm}}(\psi^{(j)}_t) \times S_{\text{geo}}(\psi^{(j)}_t)$
        \If{$\psi^{(j)}_t$ is not OOD} \Comment{As per Eq.~\ref{eq:ood-detection}}
            \State $C \gets C \cup \{j\}$ \Comment{Add to candidate set}
        \EndIf
    \EndFor
    \State $j^* = \argmax_{j \in C}\; S_{\text{skill}}(\psi^{(j)}_t)$
    \State \Return $\psi^{(j^*)}_{1:t}, s^{(j^*)}_{1:t+1}$ \Comment{Return running plan}
\EndFunction
\end{algorithmic}
}
\end{algorithm}

\textbf{Skill feasibility:} 
The second factor of Eq.~\ref{eq:tamp-step-score-factor} captures the \textit{feasibility} of a skill generated by the LLM. 
We define the skill feasibility score $S_{\text{geo}}$:
\begin{align}
    S_{\text{geo}}(\psi_t) 
        &= p(r_t \mid i, s_1, \psi_{1:t}, r_{1:t-1}) \label{eq:motion-step-score} \\
        &\approx Q^{\pi_t}(s_t, a^*_t) \label{eq:motion-step-score-decomp},
\end{align} 
where Eq.~\ref{eq:motion-step-score-decomp} approximates Eq.~\ref{eq:motion-step-score} by the Q-value evaluated at predicted future state $s_t$ with optimized parameter $a^*_t$, both of which are computed by the geometric feasibility planner.
We refer to Appendix~\ref{appx-sub:skill-feasibility} for a detailed derivation of Eq.~\ref{eq:motion-step-score-decomp}.

\textbf{Skill selection:}
The skill feasibility score (Eq.~\ref{eq:motion-step-score-decomp}) and skill usefulness score (Eq.~\ref{eq:llm-step-score-decomp}) are then multiplied to produce the overall skill score (Eq.~\ref{eq:tamp-step-score-factor}) for each of the $K$ candidate skills $\{\psi_t^1, \dots, \psi_t^K\}$. 
Invalid skills as determined by Section~\ref{subsec:ood-detection} are filtered-out of the candidate set.
Of the remaining skills, the one with the highest skill score $\psi^*_t$ is added to the running plan $\psi_{1:t-1}$.
If the predicted geometric state $s_{t+1}$ that results from skill $\psi^*_t$ satisfies the predicted goal condition (Eq.~\ref{eq:satisfaction-function}), the skill sequence $\psi_{1:t}$ is returned for execution.
Otherwise, $s_{t+1}$ is used to initialize planning iteration $t+1$.
The process repeats until the planner returns or a maximum search depth $d_{\text{max}}$ is met raising a \texttt{planning failure}.
This process is outlined in Algorithm~\ref{alg:greedy-search}.

The baselines we compare to~\cite{saycan-2022, innermono-2022} only consider the feasibility of skills $\psi^k_t$ in the current state $s_t$. 
In contrast, \se{} considers the feasibility of skills $\psi^k_t$ in the context of the planned sequence $\psi_{1:t-1}$ via geometric feasibility planning.

\begin{figure}
    \centering    
    \includegraphics[width=0.50\textwidth]{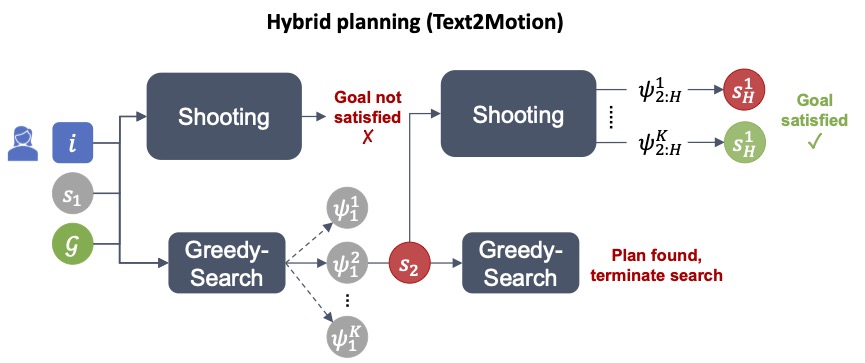}
    \caption{
    \updated{
    \textbf{Proposed hybrid planner}. After predicting goals for a given instruction, \ttmnb{} iterates the process: i) invoke \shnb{} to plan full skill sequences, and if no goal-reaching plan is found, ii) take a \senb{} step and check if executing the selected “best” skill would reach the goal.
    Note that the entire planning process occurs before execution. See Figure~\ref{fig:system} for a visualization of the \shnb{} and \senb{} planners.
    }
    }
    \label{fig:overall-method}
\end{figure}

\subsection{\ttm{}}
\label{subsec:text2motion}
We present \ttm{}, a hybrid planning algorithm that inherits the strengths of both shooting-based and search-based planning strategies.
In particular, \sh{} offers efficiency when geometrically feasible skill sequences can be easily predicted by the LLM given the initial state and the instruction. 
\se{} serves as a reliable fall-back strategy that can determine what skills are feasible at the current timestep, should \sh{} fail to find a plan.
A visualization is provided in Figure~\ref{fig:overall-method}.

At each planning iteration $t$, \ttm{} optimistically invokes \sh{} to plan $K$ candidate skill sequences.
If \sh{} raises a \texttt{planning failure}, then \ttm{} falls back to a single step of \se{}, which adds the skill $\psi^*_t$ with the highest skill score (Eq.~\ref{eq:tamp-step-score-factor}) to the running plan $\psi_{1:t-1}$.
The geometric feasibility planner predicts the state $s_{t+1}$ that would result from executing $\psi^*_t$.
If state $s_{t+1}$ satisfies the goal condition (Eq.~\ref{eq:satisfaction-function}), the skill sequence $\psi_{1:t}$ is returned for execution.
Otherwise, the next planning iteration starts by invoking \sh{} on predicted state $s_{t+1}$.
The process repeats until the planner returns or a maximum search depth $d_{\text{max}}$ is met.
\ttm{} is outlined in Algorithm~\ref{alg:text2motion}.

\begin{algorithm}[t]
\updated{
\algblock[TryCatchFinally]{try}{endtry}
\algcblock[TryCatchFinally]{TryCatchFinally}{finally}{endtry}
\algcblockdefx[TryCatchFinally]{TryCatchFinally}{catch}{endtry}
	[1]{\textbf{catch} #1}
	{\textbf{end try}}
\small
\caption{Text2Motion hybrid planner}\label{alg:text2motion}
\begin{algorithmic}[1]
\State \textbf{globals:} $\mathcal{L}^\chi, \textsc{SatFunc}, \textsc{Shooting}, \textsc{Greedy-Step}$
\Function{Text2Motion}{$i, s_1, \mathcal{G}; K, d_{\text{max}}$}
    \State $F^\mathcal{G}_{\text{sat}} \gets \Call{SatFunc}{\mathcal{G}, \mathcal{L}^\chi}$ \Comment{Goal checker}
    \State $\Psi = [\,]$; $\tau = [s_1]$ \Comment{Init. running plan}
    \While{$len(\Psi) < d_{\text{max}}$}
        \try
            \State \Return $\Call{Shooting}{i, \tau, \mathcal{G}, K}$
        \catch{\texttt{planning failure}}
            \State $\Psi, \tau \gets \Call{Greedy-Step}{i, s_1, \mathcal{G}, \Psi, \tau, K}$
            \If{$F^\mathcal{G}_{\text{sat}}(\tau[-1]) == 1$} 
                \State \Return $\Psi$
            \EndIf
        \endtry
    \EndWhile
    \State \textbf{raise} \texttt{planning failure}
\EndFunction
\end{algorithmic}
}
\end{algorithm}

\subsection{Out-of-distribution detection}
\label{subsec:ood-detection}
During planning, the LLM may propose skills that are out-of-distribution (OOD) given a state $s_t$ and optimized parameter $a^*_t$.
For instance, a symbolically incorrect skill, like \graytext{Place(dish, table)} when the \graytext{dish} is not in hand, may end up being selected if we rely on learned Q-values, since the Q-value for an OOD input can be spuriously high. 
We therefore reject plans that contain an OOD skill.

We consider a skill $\psi_t$ to be OOD if the variance of its Q-value (Eq.~\ref{eq:motion-step-score-decomp}) predicted by an ensemble~\cite{lakshminarayanan2017simple} exceeds a calibrated threshold $\epsilon^{\psi_t}$:
\begin{equation}
    \label{eq:ood-detection}
    \func{F_{\text{OOD}}}{\psi_t} = \mathbbm{1} \left(\V{i\sim1:B}{Q^{\pi_t}_i(s_t, a_t^*)} \geq \epsilon^{\psi_t} \right),
\end{equation}
where $\mathbbm{1}$ is the indicator function and $B$ is the ensemble size. 
\rtwo{We refer to Appendix~\ref{appx-sub:ood-calibration} for details on calibrating OOD thresholds $\epsilon^{\psi}$.}

\section{Experiments}
\label{sec:experiments}
We conduct experiments to test four hypotheses:
\begin{description}
    \item[\textbf{H1}] Geometric feasibility planning is a necessary ingredient when using LLMs and robot skills to solve manipulation tasks with geometric dependencies from a natural language instruction.
    \item[\textbf{H2}] \se{} is better equipped to solve tasks with partial affordance perception (as defined in Section~\ref{subsec:task-suite}) compared to \sh{}.
    \item[\textbf{H3}] \ttm{}'s hybrid planner inherits the strengths of shooting- and search-based strategies.
    \item[\textbf{H4}] \textit{A priori} goal prediction is a more reliable plan termination strategy than \texttt{stop} scoring.
\end{description}

The following subsections describe the baseline methods we compare against, details on LLMs and prompts, the tasks over which planners are evaluated, and performance metrics we report.

\subsection{Baselines}
\label{subsec:baselines}
We compare \ttm{} with a series of language-based planners, including the proposed \sh{} and \se{} strategies.
For consistency, we use the same skill library $\mathcal{L^\psi}$, with independently trained policies $\pi$ and Q-functions $Q^{\pi}$, the OOD rejection strategy (Section~\ref{subsec:ood-detection}) and, where appropriate, the dynamics models $T^\pi(s, a)$ and geometric feasibility planner (Section~\ref{subsec:geometric-feasibility-planning}) across all methods and tasks.

\scgs{}:
We implement a cost-considerate variant of SayCan~\cite{saycan-2022} with a module dubbed \gs{} (GS).
At each timestep $t$, SayCan ranks \textit{all possible} skills by $p(\psi_t \mid i, \psi_{1:t-1}) \cdot V^{\pi_t}(s_t)$, before executing the top scoring skill (Scorer).
However, the cost of ranking skills scales unfavorably with the number of scene objects $\mathcal{O}$ and skills in library $\mathcal{L}^\psi$.
\scgs{} limits the pool of skills considered in the ranking process by querying the LLM for the $K$ most useful skills $\{\psi_t^1, \dots, \psi_t^K\} \sim p(\psi_t \mid i, \psi_{1:t-1})$ (Generator) before engaging Scorer.
Execution terminates when the score of the \texttt{stop} ``skill'' is larger than the other skills.

\imgs{}:
We implement the \textit{Object + Scene} variant of Inner Monologue~\cite{innermono-2022} by providing task-progress scene context in the form of the environment's symbolic state.
We acquire \imgs{} by equipping~\cite{innermono-2022} with \gs{} for cost efficiency. 
LLM skill likelihoods are equivalent to those from \scgs{} except they are now also conditioned on the visited state history $p(\psi_t \mid i, s_{1:t}, \psi_{1:t-1})$.

\subsection{Large language model}
\label{subsec:llm}
We use two pretrained language models, both of which were accessed through the OpenAI API: i) \texttt{text-davinci-003}, a variant of the InstructGPT \cite{ouyang2022training} language model family which is finetuned from GPT-3 with human feedback and ii) the Codex model \cite{chen2021evaluating} (specifically, \texttt{code-davinci-002)}. 
For the \sh{} planner, we empirically found \texttt{text-davinci-003} to be the most capable at open-ended generation of skill sequences.
For all other queries, we use \texttt{code-davinci-002} as it was found to be reliable. 
We do not train or finetune the LLMs and only use few shot prompting.

\subsection{Prompt engineering}
\label{subsec:prompt-engineering}
The in-context examples are held consistent across all methods and tasks in the prompts passed to the LLM. 
We provide an example of the prompt structure used to query \se{} for $K=5$ skills at the first planning iteration (prompt template is in \texttt{black} and LLM output is in \texttt{\textcolor{orange}{orange}}):

\vspace{0.3cm}
\noindent\fbox{\parbox{0.97\linewidth}{\small{\texttt{{
Available scene objects: [`table', `hook', `rack', `yellow box', `blue box', `red box']\\\\
Object relationships: [`inhand(hook)', `on(yellow box, table)', `on(rack, table)', `on(blue box, table)']\\\\
Human instruction: How would you push two of the boxes to be under the rack?\\\\
Goal predicate set: {\color{code-constant}[[`under(yellow box, rack)', `under(blue box, rack)'], [`under(blue box, rack)', `under(red box, rack)'], [`under(yellow box, rack)', `under(red box, rack)']]}\\\\
Top 5 next valid robot actions (python list): {\color{code-constant}['push(yellow box, rack)', 'push(red box, rack)', 'place(hook, table)', 'place(hook, rack)', 'pull(red box, hook)']}
}}}}}
\vspace{0.3cm}

\noindent
The prompt above chains the output of two queries together: one for goal prediction (Section~\ref{subsec:goal-prediction}), and another for skill generation (Section~\ref{sec:greedy-search}). 
To compute the skill usefulness (Eq.~\ref{eq:llm-step-score-decomp}), we replace \texttt{Top 5 next valid robot actions} with \texttt{Executed action:}, append the language description of the generated skill (e.g. \graytext{Push(yellow box, rack)}), and sum token log-probabilities. 
We provide the full set of in-context examples in the Appendix (Appendix~\ref{appx-sub:suppl-incontext-examples}).

\begin{figure}
    \centering    
    \includegraphics[width=0.46\textwidth]{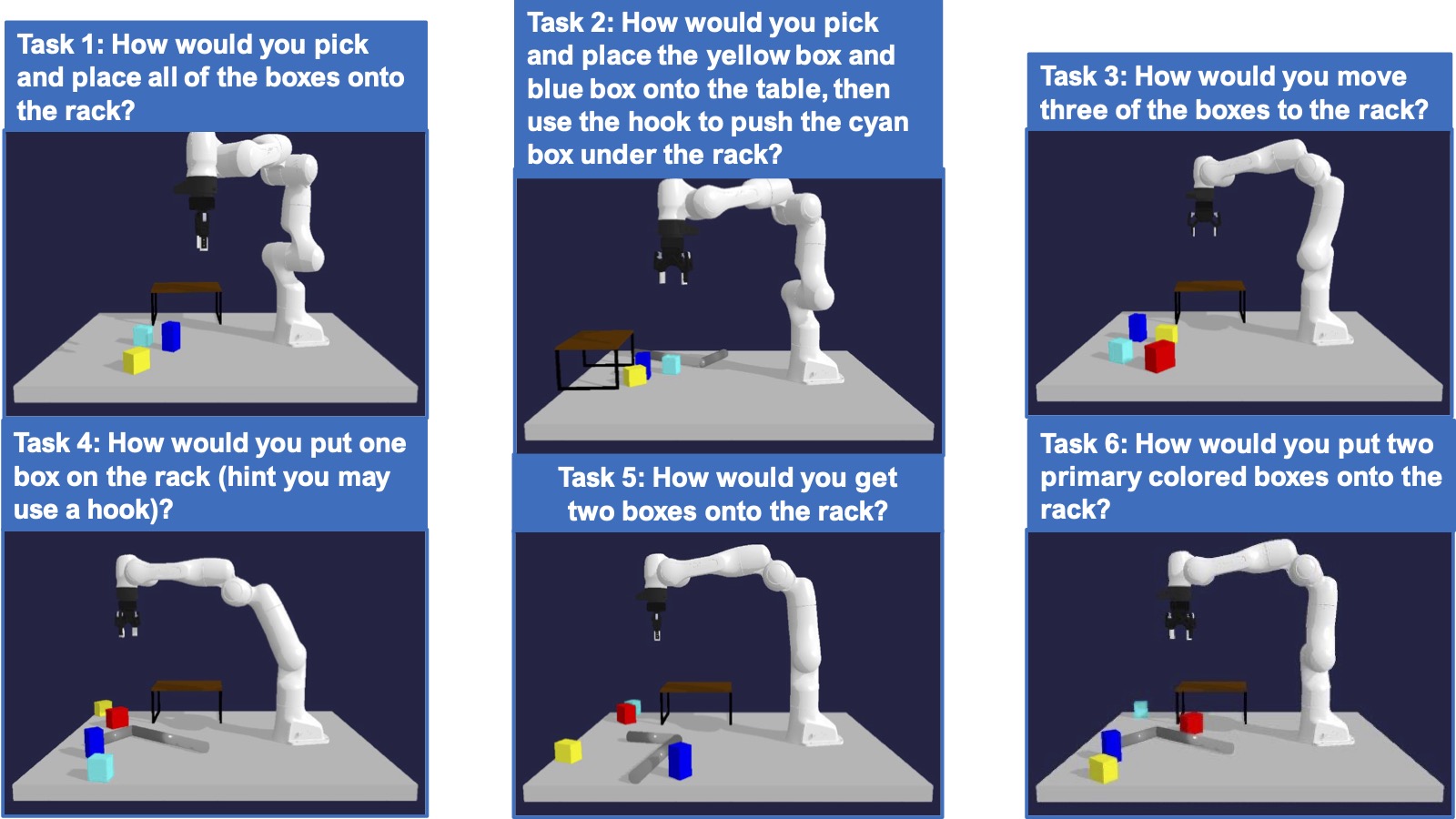}
    \vspace{10pt}
    \caption{\textbf{TableEnv Manipulation evaluation task suite}. We evaluate the performance of all methods on tasks based on the above manipulation domain. The tasks considered vary in terms of difficulty and each task contains a subset of three properties: being long horizon (Tasks 1, 2, 3, 5, 6), containing lifted goals (Tasks 4, 5, 6), and having partial affordance perception (Tasks 4, 5, 6). During evaluation, we randomize the geometric parameters of each task.}
    \label{fig:evaluation_task_suite}
\end{figure}

\subsection{Task suite}
\label{subsec:task-suite}
We construct a suite of evaluation tasks (Figure~\ref{fig:evaluation_task_suite}) in a table-top manipulation domain.
Each task includes a natural language instruction $i$ and initial state distribution $\rho(s)$ from which geometric task instances are sampled. 
For the purpose of experimental evaluation only, tasks also contain a ground-truth goal criterion to evaluate whether a plan has satisfied the corresponding task instruction.
Finally, each task contains subsets of the following properties:
\begin{itemize}
    \item \textbf{Long-horizon (LH):} Tasks that require skill sequences $\psi_{1:H}$ of length six or greater to solve. For example, Task 1 in Figure~\ref{fig:evaluation_task_suite} requires the robot to pick and place three objects for a total of six skills. In our task suite, \textbf{LH} tasks also contain geometric dependencies that span across the sequence of skills which are unlikely to be resolved by myopically executing each skill. For example, Task 2 (Figure~\ref{fig:evaluation_task_suite}) requires the robot to pick and place obstructing boxes (i.e. blue and yellow) to enable a collision-free push of the cyan box underneath the rack using the hook.
    \item \textbf{Lifted goals (LG):} Goals are expressed over object classes rather than object instances. For example, the lifted goal instruction ``move three boxes to the rack'' specifies an object class (i.e. boxes) rather than an object instance (e.g. the red box). This instruction is used for Task 3 (Figure ~\ref{fig:evaluation_task_suite}). 
    Moreover, \LG{} tends to correspond to planning tasks with many possible solutions. For instance, there may only be a single solution to the non-lifted instruction ``fetch me the red box and the blue box,'' but an LLM must contend with more options when asked to, for example, ``fetch any two boxes.''
    \item \textbf{Partial affordance perception (PAP):} Skill affordances cannot be perceived solely from the spatial relations described in the initial state $s_1$. For instance, Task 5 (Figure~\ref{fig:evaluation_task_suite}) requires the robot to put two boxes onto the rack. However, the scene description obtained through predicate classifiers $\mathcal{L}^\chi$ (described in Section~\ref{subsec:goal-prediction}) and the instruction $i$ do not indicate whether it is necessary to use a hook to pull an object closer to the robot first. 
\end{itemize}

\subsection{Evaluation and metrics}
\label{subsec:evaluation-metrics}
\ttm{}, \sh{}, \se{}:
We evaluate these language planners by marking a plan as successful if, upon execution, they reach a final state $s_{H+1}$ that satisfies the instruction $i$ of a given task. 
A plan is executed only if the geometric feasibility planner predicts a state that satisfies the inferred goal conditions (Section~\ref{subsec:goal-prediction}).

Two failure cases are tracked: i) planning failure: the method does not produce a sequence of skills $\psi_{1:H}$ whose optimized parameters $a^*_{1:H}$ (Eq.~\ref{eq:taps-objective}) results in a state that satisfies $F^\mathcal{G}_{\text{sat}}$ within a maximum plan length of $d_{\text{max}}$; ii) execution failure: the execution of a plan that satisfies $F^\mathcal{G}_{\text{sat}}$ does not achieve the ground-truth goal of the task.

Since the proposed language planners use learned dynamics models to optimize parameters $a_{1:H}$ with respect to (potentially erroneous) future state predictions $s_{2:H}$, we perform the low-level execution of the skill sequence $\psi_{1:H}$ in closed-loop fashion. 
Thus, upon executing the skill $\psi_{t}$ at timestep $t$ and receiving environment feedback $s_{t+1}$, we call STAP~\cite{taps-2022} to perform geometric feasibility planning on the remaining planned skills $\psi_{t+1:H}$. 
We do not perform task-level replanning, which would involve querying the LLM at timestep $t+1$ for a new sequence of skills $\psi_{t+1:H}$.

\scgs{} \textbf{\&} \imgs{}:
These myopic agents execute the next best admissible skill $\psi_t$ at each timestep $t$ without looking-ahead. 
Hence, we evaluate them in a closed-loop manner for a maximum of $d_{\text{max}}$ steps.
We mark a run as a success if the agent issues the \texttt{stop} skill and the current state $s_t$ satisfies the ground-truth goal. 
Note that this comparison is advantageous for these myopic agents because they are given the opportunity to perform closed-loop replanning at the task-level (e.g. re-attempting a failed skill), whereas task-level replanning does not occur for \ttm{}, \sh{}, or \se{}.
This advantage does not lead to measurable performance gains on the challenging evaluation domains that we consider.

\textbf{Reported metrics:} 
We report success rates and subgoal completion rates for all methods.
Success rates are averaged over ten random seeds per task, where each seed corresponds to a different geometric instantiation of the task (Section~\ref{subsec:task-suite}).
Subgoal completion rates are computed over all plans by measuring the number of steps an \textit{oracle planner} would take to reach the ground-truth goal from a planner's final state.
To further delineate the performance of \ttm{} from \sh{} and \se{}, we also report the percentages of planning and execution failures.

\section{Results}
\label{sec:results}

\begin{figure}
    \centering
    \includegraphics[width=0.48\textwidth]{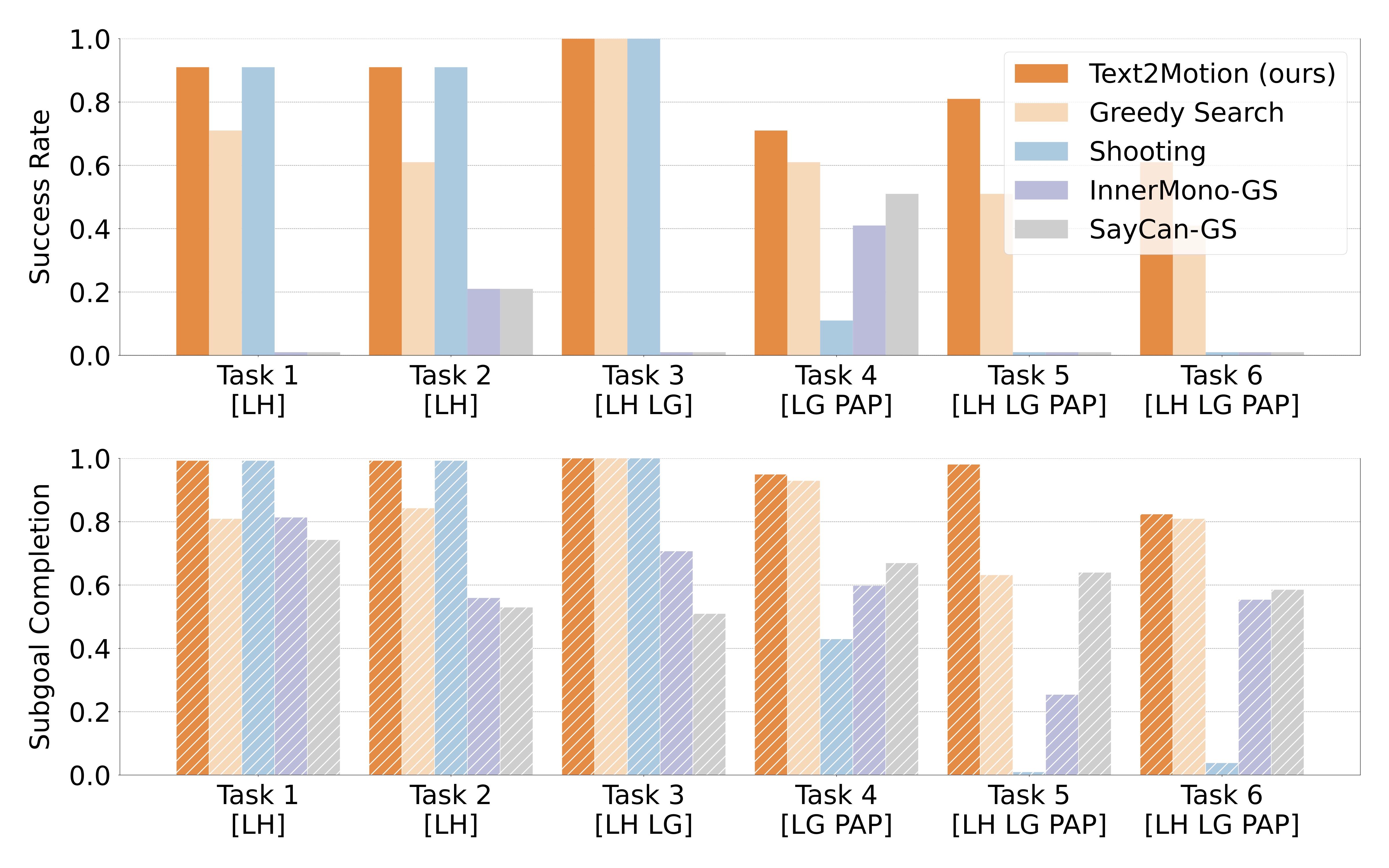}
    \vspace{-10pt}
    \caption{
    \textbf{Results on the TableEnv manipulation domain} with 10 random seeds for each task. {\bf Top:} Our method (Text2Motion) significantly outperforms all baselines on tasks involving partial affordance perception (Task 4, 5, 6). For tasks without partial affordance perception, the methods that use geometric feasibility planning (\ttmnb{}, \shnb{}, \senb{}) convincingly outperform the methods (\scgsnb~and \imgsnb) that do not. We note that \shnb{}~performs well on the tasks without partial affordance perception as it has the advantage of outputting \textit{multiple} goal-reaching candidate plans and selecting the one with the highest execution success probability. {\bf Bottom:} Methods without geometric feasibility planning tend to have high sub-goal completion rates but very low success rates. This divergence arises because it is possible to make progress on tasks without resolving geometric dependencies in the earlier timesteps; however, failure to account for geometric dependencies results in failure of the overall task. }
    \label{fig:planning-result}
\end{figure}

\subsection{Feasibility planning is required to solve tasks with geometric dependencies (\textbf{H1})}
\label{subsec:components-tamp}

Our first hypothesis is that performing geometric feasibility planning on task plans output by the LLM is essential to task success. To test this hypothesis, we compare methods that use geometric feasibility planning (\ttm{}, \sh{}, \se{}) against myopic methods that do not (\scgs{} and \imgs{}).

Instructions $i$ provided in the first two planning tasks (\LH{}) allude to skill sequences that, if executed appropriately, would solve the task.
In effect, the LLM plays a lesser role in contributing to plan success, as its probabilities are conditioned to mimic the skill sequences in $i$.
On such tasks, \ttm{}, \sh{} and \se{} which employ geometric feasibility planning over skills sequences better contend with geometric dependencies prevalent in \LH{} tasks and thereby demonstrate higher success rates. 

In contrast, the myopic baselines (\scgs{} and \imgs{}) fail to surpass success rates of 20\%, despite completing between 50\%-80\% of the subgoals (Figure~\ref{fig:planning-result}).
This result is anticipated as the feasibility of downstream skills requires coordination with earlier skills in the sequence, which these methods do not consider.
As the other tasks combine aspects of \LH{} with \LG{} and \PAP{}, it remains difficult for \scgs{} and \imgs{} to find solutions.

Surprisingly, we see that \scgs{} closely matches the performance of \imgs{}, which is additionally provided with descriptions of all states encountered during execution (as opposed to just the initial scene description).
This result suggests that explicit language feedback does not contribute to success on our tasks when considered in isolation from plan feasibility.

\begin{figure}
    \centering    
    \includegraphics[width=0.48\textwidth]{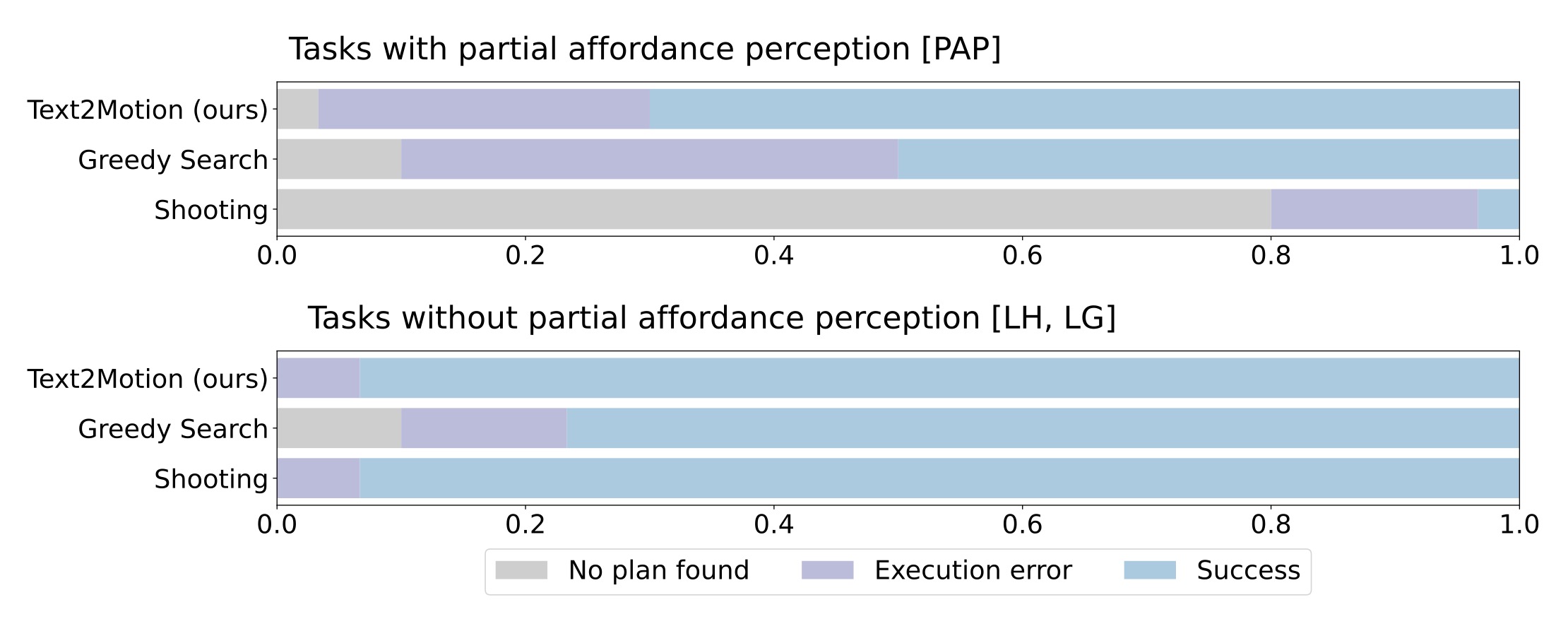}
    \vspace{-10pt}
    \caption{\textbf{Failure modes of language-based planners on two categories of tasks}.
    In this plot, we analyse the various types of failure modes that occur with \ttmnb{}, \shnb{} and \senb{} when evaluated on tasks with partial affordance perception (PAP; see Section~\ref{subsec:task-suite} for an explanation) and tasks without partial affordance perception (non-PAP). 
    \textbf{Top}: For the PAP tasks, \shnb{} incurs many planning failures because the space of possible plans is large but only few can be feasibly executed. In contrast, \senb{} uses value functions during search to narrow down the space of plans to those that are feasible.
    \ttmnb{} relies on \senb{} as a fallback if \shnb{} fails, and thus can also contend with PAP tasks. 
    \textbf{Bottom}: For the non-PAP tasks, \shnb{} outperforms \senb{}. We attribute this difference to \shnb{}'s ability to output multiple task plans while \senb{} can only output a single plan. Finally, \ttmnb{} matches the performance of \shnb{} as it also outputs and selects among multiple task plans.}
    \label{fig:failure}
\end{figure}

\subsection{Search-based reasoning is required for PAP tasks
 (\textbf{H2})}
\label{subsec:integrated-planning}
Our second hypothesis is that search-based reasoning is required to solve the \PAP{} family of tasks (defined in Section~\ref{subsec:task-suite}). 
We test this hypothesis by comparing \se{} and \sh{}, which represent two distinct approaches to combining symbolic and geometric reasoning to maximize the overall planning objective (Eq.~\ref{eq:tamp-score}).
\sh{} uses Q-functions of skills to optimize $K$ skill sequences (Eq.~\ref{eq:taps-objective}) \textit{after} they are generated by the LLM.
\se{} uses Q-functions as skill feasibility heuristics (Eq.~\ref{eq:motion-step-score-decomp}) to guide search \textit{while} a skill sequence is being constructed.

In the first two tasks (\LH{}, Figure~\ref{fig:planning-result}), we find that \sh{} achieves slightly higher success rates than \se{}, while both methods achieve 100\% success rates in the third task (\LH{} + \LG{}).
This result indicates a subtle advantage of \sh{} when multiple feasible plans can be directly inferred from $i$ and $s_1$. \sh{} can capitalize on diverse orderings of $K$ generated skill sequences (including the one specified in $i$) and select the one with the highest success probability (Eq.~\ref{eq:motion-parameter-score}).
For example, Task 1 (Figure~\ref{fig:evaluation_task_suite}) asks the robot to put three boxes onto the rack; \sh{} allows the robot to test multiple different skill sequences while \se{} only outputs a single plan.
This advantage is primarily enabled by bias in the Q-functions: Eq.~\ref{eq:stap-score} may indicate that \graytext{Place(dish, rack)} then \graytext{Place(cup, rack)} is more geometrically complex than \graytext{Place(cup, rack)} then \graytext{Place(dish, rack)}, while they are geometric equivalents.

The plans considered by \se{} at planning iteration $t$ share the same sequence of predecessor skills $\psi_{1:t-1}$.
This affords limited diversity for the planner to exploit.
However, \se{} has a significant advantage when solving the \PAP{} family of problems (Figure~\ref{fig:planning-result}, Tasks 4-6).
Here, skill sequences with high success probabilities (Eq.~\ref{eq:motion-parameter-score}) are difficult to infer directly from $i$, $s_1$, and the in-context examples provided in the prompt.
As a result, \sh{} incurs an 80\% planning failure rate, while \se{} finds plans over 90\% of the time (Figure~\ref{fig:failure}).
In terms of success, \se{} solves 40\%-60\% of the \PAP{} tasks, while \sh{} achieves a 10\% success rate on Task 4 (\LG{} + \PAP{}) and fails to solve any of the latter two tasks (\LH{} + \LG{} + \PAP{}).
Moreover, \sh{} does not meaningfully advance on any subgoals, unlike \scgs{} and \imgs{}, which consider the geometric feasibility of skills at each timestep (albeit, myopically).

\subsection{Hybrid planning integrates the strengths of shooting-based and search-based methods (\textbf{H3})}
\label{subsec:hybrid-planning}
Our third hypothesis is that shooting-based planning and search-based planning have complementing strengths that can be unified in a hybrid planning framework.
We test this hypothesis by comparing the performance of \ttm{} against \sh{} and \se{}. 

The results are presented in Figure~\ref{fig:planning-result}.
We find that \ttm{} matches the performance of \sh{} on tasks that do not consist of \PAP{} (Task 1, 2, 3).
This is expected because \sh{} does not exhibit planning failures on these tasks (Figure~\ref{fig:failure}) and \ttm{} starts by invoking \sh{}, which results in their identical performance.
However, on tasks with \PAP{} (Task 4, 5, 6) we observe that \ttm{} succeeds more often than \se{}.
This suggests that interleaving \sh{} and \se{} at each planning iteration enables \ttm{} to consider a more diverse set of goal-reaching solutions.
This result is corroborated in Figure~\ref{fig:failure}, where we see that \ttm{} incurs fewer planning and execution failures than \se{}.

\rtwo{In Table~\ref{tab:ablation-hybrid}, we further analyze the usage percentages of \sh{} and \se{} within successful plans executed by \ttm{}. 
The results show that, for tasks involving \PAP{}, over 90\% of solutions involve a combination of both shooting- and search-based strategies, which confirms our third hypothesis.}

\begin{table}[]
    \rtwo{
    \centering
    \resizebox{0.48\textwidth}{!}{%
    \begin{tabular}{lccc}
    \toprule
     Hybrid planning breakdown & Task 4 & Task 5 & Task 6 \\
    \midrule
    \% \shnb{} only  &  14\% &  0\% &  0\% \\
    \% \senb{} only &  0\% &  0\%  &  0\% \\
    \% Combination &  86\% & 100\%  & 100\% \\
    Avg. \senb{} Steps & 1.0  &  2.6 &  3.0 \\
    Avg. Plan Length & 5.0 &  7.0 &  7.0 \\
    \bottomrule \\
    \end{tabular}}
        \caption{
            \textbf{Ablation on hybrid planning method.} We analyze the usage percentages of both \shnb{} and \senb{} in successful plans found by our hybrid planner (see Figure~\ref{fig:overall-method}). We find that, as tasks increase in difficulty (Task 4, 5, 6), the majority of solutions involve a combination of both planners. This result indicates that shooting-based and search-based planning strategies play complementing roles in the success of \ttmnb{}.
        }
    \label{tab:ablation-hybrid}
    }
\end{table}

\subsection{Plan termination is made reliable via goal prediction (\textbf{H4})}
\label{subsec:plan-termination}
\begin{figure}
    \centering        
    \includegraphics[width=0.48\textwidth]{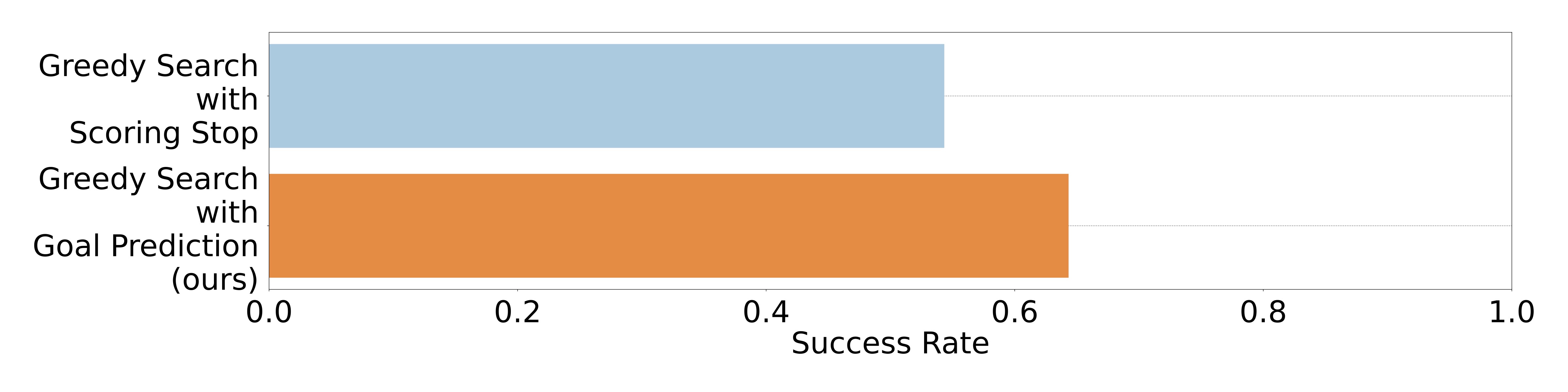}
    \vspace{-10pt}
    \caption{\textbf{Ablation on termination method: goal proposition prediction vs stop scoring}. 
    We compare the performance of \senb{} using two different plan termination methods: using the LLM to predict goals \textit{a priori} (ours) and scoring a \texttt{stop} skill~\cite{saycan-2022} during search. 
    We present results averaged across all six tasks and ten seeds for each variation (120 experiments in total). 
    We find that terminating planning when LLM-predicted goals are satisfied results in a 10\% boost in success rate over \texttt{stop} scoring.}
    \label{fig:ablation-termination}
\end{figure}

Our fourth hypothesis is that predicting goals from instructions \textit{a priori} and selecting plans based on their satisfication (Section~\ref{subsec:goal-prediction}) is more reliable than 
scoring plan termination with a dedicated \texttt{stop} skill at each timestep.
We test this hypothesis in an ablation experiment (Figure~\ref{fig:ablation-termination}), comparing our plan termination method to that of SayCan and Inner Monologue's, while keeping all else constant for our \se{} planner. 
We run 120 experiments (two variations, six tasks, and ten seeds each) in total on the TableEnv Manipulation task suite. 
The results in Figure~\ref{fig:ablation-termination} suggest that, for the tasks we consider, our proposed goal prediction method leads to 10\% higher success rates than the scoring baseline. 

We also note the apparent advantages of both techniques.
First, goal prediction is more efficient than scoring \texttt{stop} as the former requires only one LLM query, whereas the latter needs to be queried at every timestep.
Second, goal prediction offers interpretability over \texttt{stop} scoring, as it is possible to inspect the goal that the planner is aiming towards prior to execution. 
Nonetheless, \texttt{stop} scoring does provide benefits in terms of expressiveness, as its predictions are not constrained to any specific output format.
This advantage, however, is not captured in our evaluation task suite, which at most require conjunctive ($\land$) and disjunctive ($\lor$) goals.
For instance, ``Stock two boxes onto the rack'' could correspond to (\graytext{on(red box, rack)} $\land$ \graytext{on(blue box, rack)}) $\lor$ (\graytext{on(yellow box, rack)} $\land$ \graytext{on(cyan box, rack)}), while in theory, \texttt{stop} scoring can represent all goals expressible in language.

\section{Limitations and Future Work}
\label{sec:limitations}
\rtwo{
\textbf{LLM likelihoods:} 
We observed an undesirable pattern emerge in the planning phase of \se{} and the execution phase of \scgs{} and \imgs{}, where \textit{recency bias}~\cite{zhao2021calibrate} would cause the LLM to produce unreliable likelihoods (Eq.~\ref{eq:llm-step-score-decomp}), inducing a cyclic state of repeating feasible skills.
While we mitigate such failures by combining \se{} and \sh{} in the hybrid \ttm{} algorithm, leveraging calibration techniques to increase the reliability LLM likelihoods~\cite{li2022contrastive, chen2022close} may improve the performance search-based planning over long-horizons.}

\textbf{Skill library:} 
As with other methods that use skill libraries, \ttm{} is reliant on the fidelity of the learned skills, their value functions, and the ability to accurately predicted future states with dynamics models.
\rtwo{Thus, incorporating skills that operate on high-dimensional observations (e.g. images~\cite{concept2robot-2021}) into our framework may require adopting techniques for stable long-term predictions in these observation spaces~\cite{2022-driess-compNerf}}.

\textbf{Runtime complexity:} 
\ttm{} is mainly comprised of learned components, which comes with its associated efficiency benefits. Nonetheless, runtime complexity was not a core focus of this work, and STAP~\cite{taps-2022} was frequently called during planning, increasing the overall planning time.
LLMs also impose an inference bottleneck as each API query (Section \ref{subsec:llm}) requires 2-10 seconds, but we anticipate improvements with advances in both LLM inference techniques and in methods that distill LLM capabilities into smaller, cost-efficient models~\cite{touvron2023llama}.

\rtwo{\textbf{Closed-world assumptions:} Our framework operates in a closed-world setting (Section~\ref{sec:problem-setup}), where we assume to know which objects are task-relevant and the poses of objects are estimated by an external perception system at the time of receiving a language instruction.} \rone{Extending \ttm{} to open-world settings may necessitate exploring~\cite{chen2022open} or interacting~\cite{curtis2022-unknown} with the environment to discover objects that are not initially observable, and training skills to support a diverse set of real-world objects~\cite{jang2021bczero, mtopt2021arxiv}.}

\textbf{Future Work:} We outline several avenues for future work based on these observations.
First, there remains an opportunity to increase the plan-time efficiency of our method, for instance, by warm starting geometric feasibility planning with solutions cached in earlier planning iterations~\cite{williams2015model}.
\rtwo{Second, we aim to explore the use of Visual Question and Answering (VQA)~\cite{zhou2020unified} and multi-modal foundation models that are visually grounded~\cite{driess2023palm, openai2023gpt4}. 
Such models may support scaling \ttm{} to higher-dimensional observation spaces and potentially serve as a substitutes for closed-world components used in our framework (e.g. detecting a variable number of predicates using VQA).}
\rtwo{Lastly, we hope to leverage \ttm{} as part of a broader planning system enroute to the goal of open-world operability. Such a system could, for example, use Text2Motion to produce feasible and verified plans to subgoals, while building knowledge of the environment in unobserved or partially observable regions during the execution of those subgoals.}
Progress on each of these fronts would constitute steps in the direction of scalable, reliable, and real-time language planning capabilities.

\section{Conclusion}
\label{sec:conclusion}

We present a language-based planning framework that combines LLMs, learned skills, and geometric feasibility planning to solve long-horizon robotic manipulation tasks containing geometric dependencies.
\ttm{} constructs a task- and motion-level plan and verifies that it satisfies a natural language instruction by testing planned states against inferred goals.
In contrast to prior language planners, our method verifies that its plan satisfies the instruction before executing any actions in the environment.
\ttm{} represents a hybrid planning formalism that optimistically queries an LLM for long-horizon plans and falls back to a reliable search strategy should optimistic planning fail. 
As a result, \ttm{} inherits the strengths of both shooting-based and search-based planning formalisms.

Our results highlight the following: (i) geometric feasibility planning is important when using LLMs and learned skills to solve sequential manipulation tasks from natural language instructions; (ii) search-based reasoning can contend with a family of tasks where the space of possible plans is large but only few are feasible; (iii) shooting-based and search-based planning strategies can be synergistically integrated in a hybrid planner that outperforms its constituent parts; (iv) terminating plans based on inferred symbolic goals is more reliable than prior LLM scoring techniques.

\backmatter

\bmhead{Acknowledgments}
Toyota Research Institute and Toshiba provided funds to support this work. 
This work was also supported by the National Aeronautics and Space Administration (NASA) under the Innovative Advanced Concepts (NIAC) program.

\bibliography{sn-bibliography}%

\clearpage
\onecolumn

\newpage
\setlength{\parskip}{1em}

\section*{Overview}
The appendix offers additional details with respect to the implementation of \ttm{} and language planning baselines (Appendix~\ref{appx:implementation-details}), the experiments conducted (Appendix~\ref{appx:experiment-details}), derivations supporting the design of our algorithms (Appendix~\ref{appx:derivations}), and the \textbf{real-world planning demonstrations} (Appendix~\ref{appx:demos}). Qualitative results are made available at \link{https://sites.google.com/stanford.edu/text2motion}{sites.google.com/stanford.edu/text2motion}.

\begin{appendices}

\startcontents[sections]
\printcontents[sections]{l}{1}{\setcounter{tocdepth}{2}}

\clearpage

\section{Implementation Details}
\label{appx:implementation-details}
The \ttm{} planner integrates both \sh{} and \se{} to construct skill sequences that are feasible for the robot to execute in the environment. 
The planning procedure relies on four core components: 1) a library of learned robot skills, 2) a method for detecting when a skill is out-of-distribution (OOD), 3) a large language model (LLM) to perform task-level planning, and 4) a geometric feasibility planner that is compatible with the learned robot skills. 
All evaluated language-based planners use the above components, while \scgs{} and \imgs{} are myopic agents that do not perform geometric feasibility planning.
We provide implementation details of these components in the following subsections.

\subsection{Learning robot skills and dynamics}
\label{appx-sub:learning-models}
\textbf{Skill library overview:} All evaluated language planners interface an LLM with a library of robot skills $\mathcal{L}=\{\psi^1, \ldots, \psi^N\}$.
Each skill $\psi$ has a language description (e.g. \graytext{Pick(a)}) and is associated with a parameterized manipulation primitive~\cite{felip2013manipulation} $\phi(a)$.
A primitive $\phi(a)$ is controllable via its \textit{parameter} $a$ which determines the motion~\cite{khatib1987unified} of the robot's end-effector through a series of waypoints.
For each skill $\psi$, we train a policy $\pi(a|s)$ to output parameters $a\in\mathcal{A}$ that maximize primitive's $\phi(a)$ probability of success in a contextual bandit setting (Eq.~\ref{eq:skill-mdp}) with a skill-specific binary reward function $R(s, a, s')$.
We also train an ensemble of Q-functions $Q_{1:B}^\pi(s, a)$ and a dynamics model $T^\pi(s' | s, a)$ for each skill, both of which are required for geometric feasibility planning. 
We discuss the calibration of Q-function ensembles for OOD detection of skills in Appendix~\ref{appx-sub:ood-calibration}.

We learn four manipulation skills to solve tasks in simulation and in the real-world: $\psi^{\text{Pick}}$, $\psi^{\text{Place}}$, $\psi^{\text{Pull}}$, $\psi^{\text{Push}}$. 
Only a single policy per skill is trained, and thus, the policy must learn to engage the primitive over objects with differing geometries (e.g. $\pi^{\text{Pick}}$ is used for both \graytext{Pick(box)} and \graytext{Pick(hook)}).
The state space $\mathcal{S}$ for each policy is defined as the concatenation of geometric state features (e.g. pose, size) of all objects in the scene, where the first $n$ object states correspond to the $n$ skill arguments and the rest are randomized.
For example, the state for the skill \graytext{Pick(hook)} would have be a vector of all objects' geometric state features with the first component of the state corresponding to the \graytext{hook}.  

\textbf{Parameterized manipulation primitives:}
We describe the parameters $a$ and reward function $R(s, a, s')$ of each parameterized manipulation primitive $\phi(a)$ below.
A collision with a non-argument object constitutes an execution failure for all skills, and as a result, the policy receives a reward of $0$.
For example, $\pi^{\text{Pick}}$ would receive a reward of $0$ if the robot collided with \graytext{box} during the execution of \graytext{Pick(hook)}. 
\begin{itemize}
    \item \graytext{Pick(obj)}: $a\sim\pi^{\text{Pick}}(a \vert s)$ denotes the grasp pose of \graytext{obj} \textit{w.r.t} the coordinate frame of \graytext{obj}. A reward of $1$ is received if the robot successfully grasps \graytext{obj}.
    \item \graytext{Place(obj, rec)}: $a\sim\pi^{\text{Place}}(a \vert s)$ denotes the placement pose of \graytext{obj} \textit{w.r.t} the coordinate frame of \graytext{rec}. A reward of $1$ is received if \graytext{obj} is stably placed on \graytext{rec}.
    \item \graytext{Pull(obj, tool)}: $a\sim\pi^{\text{Pull}}(a \vert s)$ denotes the initial position, direction, and distance of a pull on \graytext{obj} with \graytext{tool} \textit{w.r.t} the coordinate frame of \graytext{obj}. A reward of $1$ is received if \graytext{obj} moves toward the robot by a minimum of $d_{\text{Pull}}=0.05m$. 
    \item \graytext{Push(obj, tool, rec)}: $a\sim\pi^{\text{Push}}(a \vert s)$ denotes the initial position, direction, and distance of a push on \graytext{obj} with \graytext{tool} \textit{w.r.t} the coordinate frame of \graytext{obj}. A reward of $1$ is received if \graytext{obj} moves away from the robot by a minimum of $d_{\text{Push}}=0.05m$ and if \graytext{obj} ends up underneath \graytext{rec}.
\end{itemize}

\textbf{Dataset generation:} 
All planners considered in this work rely on accurate Q-functions $Q^\pi(s, a)$ to estimate the feasibility of skills proposed by the LLM. 
This places a higher fidelity requirement on the Q-functions than typically needed to learn a reliable policy, as the Q-functions must characterize both skill success (feasibility) and failure (infeasibility) at a given state.
Because the primitives $\phi(a)$ reduce the horizon of policies $\pi(a|s)$ to a single timestep, and the reward functions are $R(s, a, s')\in\{0, 1\}$, the Q-functions can be interpreted as binary classifiers of state-action pairs.
Thus, we take a staged approach to learning the Q-functions $Q^\pi$, followed by the policies $\pi$, and lastly the dynamics models $T^\pi$.

Scenes in our simulated environment are instantiated from a symbolic specification of objects and spatial relations, which together form a symbolic state.
The goal is to learn a Q-function that sufficiently covers the state-action space of each skill.
We generate a dataset that meets this requirement in four steps: a) enumerate all valid symbolic states; b) sample geometric scene instances $s$ per symbolic state; c) uniformly sample actions over the action space $a\sim\mathcal{U}^{[0,1]^d}$; (d) simulate the states and actions to acquire next states $s'$ and compute rewards $R(s, a, s')$.
We slightly modify this sampling strategy to maintain a minimum success-failure ratio of 40\%, as uniform sampling for more challenging skills such as \graytext{Pull} and \graytext{Push} seldom emits a success ($\sim$3\%).
We collect 1M $(s, a, s', r)$ tuples per skill of which 800K of them are used for training ($\mathcal{D}_t$), while the remaining 200K are used for validation ($\mathcal{D}_v$).
We use the same datasets to learn the Q-functions $Q^\pi$, policies $\pi$, and dynamics models $T^\pi$ for each skill.
 
\textbf{Model training:}
We train an ensemble of Q-functions with mini-batch gradient descent and logistic regression loss.
Once the Q-functions have converged, we distill their returns into stochastic policies $\pi$ through the maximum-entropy update~\cite{pmlr-v80-haarnoja18b}:
\begin{equation*}
    \begin{split}
        \pi^* \leftarrow \arg \max_{\pi} \, \operatorname{E}_{(s,a)\sim \mathcal{D}_t} [\min(Q_{1:B}^\pi(s, a)) \\ - \alpha \log\pi(a|s) ].
    \end{split}
\end{equation*}
Instead of evaluating the policies on $\mathcal{D}_v$, which contains states for which no feasible action exists, the policies are synchronously evaluated in an environment that exhibits only feasible states. 
This simplifies model selection and standardizes skill capabilities across primitives. 
All Q-functions achieve precision and recall rates of over 95\%. 
The average success rates of the converged policies over 100 evaluation episodes are: $\pi_{\text{Pick}}$ with 99\%, $\pi_{\text{Place}}$ with 90\%, $\pi_{\text{Pull}}$ with 86\%, $\pi_{\text{Push}}$ with 97\%.

We train a deterministic dynamics model per skill using the forward prediction loss:
\begin{equation*}
        L_{\text{dynamics}}\left(T^\pi; \mathcal{D}_t \right) = \operatorname{E}_{(s,a,s')\sim \mathcal{D}_t}||T^\pi(s, a) - s'||_2^2.
\end{equation*}
The dynamics models converge to within millimeter accuracy on the validation split.

\textbf{Hyperparameters:} The Q-functions, policies, and dynamics models are MLPs with hidden dimensions of size [256, 256] and ReLU activations.
We train an ensemble of $B=8$ Q-functions with a batch size of 128 and a learning rate of 1$e^{-4}$ with a cosine annealing decay~\cite{loshchilov2017sgdr}.
The Q-functions for \graytext{Pick}, \graytext{Pull}, and \graytext{Push} converged on $\mathcal{D}_v$ in 3M iterations, while the Q-function for \graytext{Place} required 5M iterations.
We hypothesize that this is because classifying successful placements demands carefully attending to the poses and shapes of all objects in the scene so as to avoid collisions.
The policies are trained for 250K iterations with a batch size of 128 and a learning rate of 1$e^{-4}$, leaving all other parameters the same as \cite{pmlr-v80-haarnoja18b}.
The dynamics models are trained for 750K iterations with a batch size of 512 and a learning rate of 5$e^{-4}$; only on successful transitions to avoid the noise associated with collisions and truncated episodes.
The parallelized training of all models takes approximately 12 hours on an Nvidia Quadro P5000 GPU and 2 CPUs per job.

\subsection{Out-of-distribution detection}
\label{appx-sub:ood-calibration}
The training dataset described in Section~\ref{appx-sub:learning-models} contain both successes and failures for symbolically valid skills like \graytext{Pick(box)}.
However, when using LLMs for robot task planning, it is often the case that the LLM will propose symbolically invalid skills, such as \graytext{Pick(table)}, that neither the skill's policy, Q-functions, or dynamics model have observed in training.
We found that a percentage of out-of-distribution (OOD) queries would result in erroneously high Q-values, causing the invalid skill to be selected. 
Attempting to execute such a skill leads to control exceptions or other failures.

Whilst there are many existing techniques for OOD detection of deep neural networks, we opt to detect OOD queries on the learned Q-functions via deep ensembles due to their ease of calibration~\cite{lakshminarayanan2017simple}.
A state-action pair is classified as OOD if the empirical variance of the predicted Q-values is above a determined threshold:
\begin{equation*}
    \func{F_{\text{OOD}}}{\psi} = \mathbbm{1} \left(\V{i\sim1:B}{Q^{\pi}_i(s, a)} \geq \epsilon^{\psi} \right),
\end{equation*}
where each threshold $\epsilon^\psi$ is unique to skill $\psi$.

To determine the threshold values, we generate an a calibration dataset of 100K symbolically invalid states and actions for each skill.
The process takes less than an hour on a single CPU as the actions are infeasible and need not be simulated in the environment (i.e. rewards are known to be $0$).
We compute the mean and variance of the Q-ensemble for each $(s, a)$ sample in both the training dataset (in-distribution inputs) and the calibration dataset (out-of-distribution inputs), and produce two histograms by binning the computed ensemble variances by the ensemble means. 
We observe that the histogram of variances corresponding to OOD inputs is uniform across all Q-value bins and is an order of magnitude large than the ensemble variances computed over in-distribution inputs.
This allows us to select thresholds $\epsilon^\psi$ which are low enough to reliably detect OOD inputs, yet will not be triggered for in-distribution inputs: $\epsilon^{\text{Pick}} = 0.10$, $\epsilon^{\text{Place}} = 0.12$, $\epsilon^{\text{Pull}} = 0.10$, and $\epsilon^{\text{Push}} = 0.06$.

\subsection{Task planning with LLMs}
\label{appx-sub:llm-task-planning}
\ttm{}, \se{}, and the myopic planning baselines \scgs{} and \imgs{} use \texttt{code-davinci-002}~\cite{chen2021evaluating} to generate and score skills, while \sh{} queries \texttt{text-davinci-003}~\cite{ouyang2022training} to directly output full skill sequences.
In our experiments, we used a temperature setting of $0$ for all LLM queries. 

To maintain consistency in the evaluation of various planners, we allow \ttm{}, \scgs{}, and \imgs{} to generate $K=5$ skills $\{\psi^1_t, \ldots, \psi^K_t\}$ at each timestep $t$.
Thus, every search iteration of \se{} considers five possible extensions to the current running sequence of skills $\psi_{1:t-1}$.
Similarly, \sh{} generates $K=5$ skill sequences.

As described in Section~\ref{sec:greedy-search}, skills are selected at each timestep $t$ via a combined usefulness and geometric feasibility score:
\begin{align*}
    S_{\text{skill}}(\psi_t) &= S_{\text{llm}}(\psi_t) \cdot S_{\text{geo}}(\psi_t) \\
    &\approx p(\psi_t \mid i, s_{1:t}, \psi_{1:t-1}) \cdot Q^{\pi_t}(s_t, a^*_t),
\end{align*} 
where \ttm{}, \se{}, and \sh{} use geometric feasilibity planning (details below in Appendix~\ref{appx-sub:taps-geometric-planning}) to compute $S_{\text{geo}}(\psi_t)$, while \scgs{} and \imgs{} use the current value function estimate $V^{\pi_t}(s_t) = \operatorname{E}_{a_t\sim\pi_t}[Q^{\pi_t}(s_t, a_t)]$.
We find that in both cases, taking $S_{\text{llm}}(\psi_t)$ to be the SoftMax log-probability score produces a winner-takes-all effect, causing the planner to omit highly feasible skills simply because their associated log-probability was marginally lower than the LLM-likelihood of another skill.
Thus, we dampen the SoftMax operation with a $\beta$-coefficient to balance the ranking of skills based on both feasibility and usefulness. 
We found $\beta=0.3$ to work well.

\subsection{Geometric feasibility planning}
\label{appx-sub:taps-geometric-planning}
Given a sequence of skills $\psi_{1:H}$, geometric feasibility planning computes parameters $a_{1:H}$ that maximizes the success probability of the underlying sequence of primitives $\phi_{1:H}$. 
For example, given a skill sequence \graytext{Pick(hook)}, \graytext{Pull(box, hook)}, geometric feasibility planning would compute a 3D grasp position on the hook that enables a successful pull on the box thereafter.

\ttm{} is agnostic to the method that fulfils the role of geometric feasibility planning. 
In our experiments we leverage Sequencing Task-Agnostic Policies (STAP)~\cite{taps-2022}.
Specifically, we consider the PolicyCEM variant of STAP, where optimization of the skill sequence's success probability (Eq.~\ref{eq:taps-objective}) is warm started with parameters sampled from the policies $a_{1:H}\sim \pi_{1:H}$. 
We perform ten iterations of the Cross-Entropy Method~\cite{rubinstein1999-cem}, sampling 10K trajectories at each iteration and selecting 10 elites to update the mean of the sampling distribution for the following iteration. 
The standard deviation of the sampling distribution is held constant at 0.3 for all iterations.

\clearpage

\section{Experiment Details}
\label{appx:experiment-details}

\begin{table*}[t]
    \centering
    \caption{\textbf{TableEnv manipulation task suite}. We use the following shorthands as defined in the paper: LH: Long-Horizon, LG: Lifted Goals, PAP: Partial Affordance Perception.}
    \begin{tabular}{@{}l|c|l@{}}
    \toprule
    \textbf{Task ID} & \textbf{Properties} & \textbf{Instruction}\\
    \midrule
    \textbf{Task 1} & LH & How would you pick and place all of the boxes onto the rack?” \\
    \textbf{Task 2} & LH + LG & How would you pick and place the yellow box and blue box onto the table, \\ & & then use the hook to push the cyan box under the rack?”

 \\
    \textbf{Task 3} & LH + PAP & How would you move three of the boxes to the rack?”
 \\
    \textbf{Task 4} & LG + PAP & How would you put one box on the rack?”

 \\
    \textbf{Task 5} & LH + LG + PAP & How would you get two boxes onto the rack?”

 \\
    \textbf{Task 6} & LH + LG + PAP & How would you move two primary colored boxes to the rack?”

 \\
    \bottomrule
    \end{tabular}
    \label{table:text2motion-domains}
\end{table*}

We refer to Table.~\ref{table:text2motion-domains} for an overview of the tasks in the TableEnv Manipulation suite.

\subsection{Scene descriptions as symbolic states}
\label{appx-sub:scene-descr-symbolic}
For the remainder of this section, we use the following definitions of terms:
\begin{itemize}
    \item \textbf{Predicate:} a binary-valued function over objects that evaluates to true or false (e.g. \graytext{on(a, b)})
    \item \textbf{Spatial Relation:} a predicate grounded over objects that evaluates to true (e.g. \graytext{on(rack, table)})
    \item \textbf{Predicate Classifier:} a function that implements whether a predicate is true or false in the scene. In this work, we use hand-crafted predicate classifiers for each spatial relation we model
    \item \textbf{Symbolic State:} the set of all predicates that hold true in the scene
    \item \textbf{Satisfaction Function:} a binary-valued function that takes as input a geometric state, uses the predicate classifiers to detect what predicates hold true in the geometric state, and collects those predicates into a set to form a symbolic state. The satisfaction function evaluates to true if the predicted goals (predicates) hold in the symbolic state
\end{itemize}

To provide scene context to \ttm{} and the baselines, we take a heuristic approach to converting a geometric state $s$ into a basic symbolic state.
Symbolic states consist of combinations of one or more of the following predicates: \graytext{on(a, b)}, \graytext{under(a, b)}, and \graytext{inhand(a)}. 
\graytext{inhand(a) = True} when the height of object \graytext{a} is above a predefined threshold.
\graytext{on(a, b) = True} when i) object \graytext{a} is above \graytext{b} (determined by checking if the centroid of \graytext{a}'s axis-aligned bounding box is greater than \graytext{b}'s axis-aligned bounding box), ii) \graytext{a}'s bounding box intersects \graytext{b}'s bounding box, and iii) \graytext{inhand(a) = False}. \graytext{under(a, b) = True} when \graytext{on(a, b) = False} and \graytext{a}'s bounding box intersects \graytext{b}'s bounding box.

\rone{The proposed \textbf{goal prediction method} (Section~\ref{subsec:goal-prediction}) outputs goal propositions consisting of combinations of the predicates above which have been grounded over objects (i.e. spatial relations). As an example, for the natural language instruction ``Put two of the boxes under the rack'' and a symbolic state \texttt{[on(red box, table), on(green box, rack), on(hook, rack), on(blue box, rack)]}, the LLM might predict the set of three goals \texttt{\{[under(red box, rack), under(blue box, rack)], [under(red box, rack), under(green box, rack)], [under(green box, rack), under(blue box, rack)]\}}. }
We note that objects are neither specified as within or beyond the robot workspace, as we leave it to the skill's Q-functions to determine feasibility (Section~\ref{appx-sub:learning-models}).

Since planning in high-dimensional observation spaces is not the focus of this work, we assume knowledge of objects in the scene and use hand-crafted heuristics to detect spatial relations between objects. 
There exists several techniques to convert high-dimensional observations into scene descriptions, such as the one used in \cite{zeng2022socratic}.
We leave exploration of these options to future work.

\subsection{In-context examples}
\label{appx-sub:suppl-incontext-examples}

For all experiments and methods, we use the following in-context examples to construct the prompt passed to the LLMs.

\vspace{0.3cm}
\noindent\fbox{\parbox{0.97\linewidth}{\small{\texttt{{\\Available scene objects: ['table', 'hook', 'rack', 'yellow box', 'blue box', 'red box']\\
Object relationships: ['inhand(hook)', 'on(yellow box, table)', 'on(rack, table)', 'on(blue box, table)']\\
Human instruction: How would you push two of the boxes to be under the rack?\\
Goal predicate set: [['under(yellow box, rack)', 'under(blue box, rack)'], ['under(blue box, rack)', 'under(red box, rack)'], ['under(yellow box, rack)', 'under(red box, rack)']]\\
Top 1 robot action sequences: ['push(yellow box, hook, rack)', 'push(red box, hook, rack)']\\}}}}}

\noindent\fbox{\parbox{0.97\linewidth}{\small{\texttt{{\\Available scene objects: ['table', 'cyan box', 'hook', 'blue box', 'rack', 'red box']\\
Object relationships: ['on(hook, table)', 'on(rack, table)', 'on(blue box, table)', 'on(cyan box, table)', 'on(red box, table)']\\
Human instruction: How would you push all the boxes under the rack?\\
Goal predicate set: [['under(blue box, rack)', 'under(cyan box, rack)', 'under(red box, rack)']]\\
Top 1 robot action sequences: ['pick(blue box)', 'place(blue box, table)', 'pick(hook)', 'push(cyan box, hook, rack)', 'place(hook, table)', 'pick(blue box)', 'place(blue box, table)', 'pick(hook)', 'push(blue box, hook, rack)', 'push(red box, hook, rack)']\\}}}}}

\noindent\fbox{\parbox{0.97\linewidth}{\small{\texttt{{\\Available scene objects: ['table', 'cyan box', 'red box', 'hook', 'rack']\\
Object relationships: ['on(hook, table)', 'on(rack, table)', 'on(cyan box, rack)', 'on(red box, rack)']\\
Human instruction: put the hook on the rack and stack the cyan box above the rack - thanks\\
Goal predicate set: [['on(hook, rack)', 'on(cyan box, rack)']]\\
Top 1 robot action sequences: ['pick(hook)', 'pull(cyan box, hook)', 'place(hook, rack)', 'pick(cyan box)', 'place(cyan box, rack)']\\}}}}}

\noindent\fbox{\parbox{0.97\linewidth}{\small{\texttt{{\\Available scene objects: ['table', 'rack', 'hook', 'cyan box', 'yellow box', 'red box']\\
Object relationships: ['on(yellow box, table)', 'on(rack, table)', 'on(cyan box, table)', 'on(hook, table)', 'on(red box, rack)']\\
Human instruction: Pick up any box.\\
Goal predicate set: [['inhand(yellow box)'], ['inhand(cyan box)']]\\
Top 1 robot action sequences: ['pick(yellow box)']\\}}}}}

\noindent\fbox{\parbox{0.97\linewidth}{\small{\texttt{{\\Available scene objects: ['table', 'blue box', 'cyan box', 'hook', 'rack', 'red box', 'yellow box']\\
Object relationships: ['inhand(hook)', 'on(red box, rack)', 'on(yellow box, table)', 'on(blue box, table)', 'on(cyan box, rack)', 'on(rack, table)']\\
Human instruction: could you move all the boxes onto the rack?\\
Goal predicate set: [['on(yellow box, rack)', 'on(blue box, rack)']]\\
Top 1 robot action sequences: ['pull(yellow box, hook)', 'place(hook, table)', 'pick(yellow box)', 'place(yellow box, rack)', 'pick(blue box)', 'place(blue box, rack)']\\}}}}}

\noindent\fbox{\parbox{0.97\linewidth}{\small{\texttt{{\\Available scene objects: ['table', 'blue box', 'red box', 'hook', 'rack', 'yellow box']\\
Object relationships: ['on(hook, table)', 'on(blue box, table)', 'on(rack, table)', 'on(red box, table)', 'on(yellow box, table)']\\
Human instruction: situate an odd number greater than 1 of the boxes above the rack\\
Goal predicate set: [['on(blue box, rack)', 'on(red box, rack)', 'on(yellow box, rack)']]\\
Top 1 robot action sequences: ['pick(hook)', 'pull(blue box, hook)', 'place(hook, table)', 'pick(blue box)', 'place(blue box, rack)', 'pick(red box)', 'place(red box, rack)', 'pick(yellow box)', 'place(yellow box, rack)']\\}}}}}

\noindent\fbox{\parbox{0.97\linewidth}{\small{\texttt{{\\Available scene objects: ['table', 'cyan box', 'hook', 'red box', 'yellow box', 'rack', 'blue box']\\
Object relationships: ['on(hook, table)', 'on(red box, table)', 'on(blue box, table)', 'on(cyan box, table)', 'on(rack, table)', 'under(yellow box, rack)']\\
Human instruction: How would you get the cyan box under the rack and then ensure the hook is on the table?\\
Goal predicate set: [['under(cyan box, rack)', 'on(hook, table)']]\\
Top 1 robot action sequences: ['pick(blue box)', 'place(blue box, table)', 'pick(red box)', 'place(red box, table)', 'pick(hook)', 'push(cyan box, hook, rack)', 'place(hook, table)']\\}}}}}

\noindent\fbox{\parbox{0.97\linewidth}{\small{\texttt{{\\Available scene objects: ['table', 'cyan box', 'hook', 'yellow box', 'blue box', 'rack']\\
Object relationships: ['on(hook, table)', 'on(yellow box, rack)', 'on(rack, table)', 'on(cyan box, rack)']\\
Human instruction: set the hook on the rack and stack the yellow box onto the table and set the cyan box on the rack\\
Goal predicate set: [['on(hook, rack)', 'on(yellow box, table)', 'on(cyan box, rack)']]\\
Top 1 robot action sequences: ['pick(yellow box)', 'place(yellow box, table)', 'pick(hook)', 'pull(yellow box, hook)', 'place(hook, table)']\\}}}}}

\noindent\fbox{\parbox{0.97\linewidth}{\small{\texttt{{\\Available scene objects: ['table', 'cyan box', 'hook', 'rack', 'red box', 'blue box']\\
Object relationships: ['on(hook, table)', 'on(blue box, rack)', 'on(cyan box, table)', 'on(red box, table)', 'on(rack, table)']\\
Human instruction: Move the warm colored box to be underneath the rack.\\
Goal predicate set: [['under(red box, rack)']]\\
Top 1 robot action sequences: ['pick(blue box)', 'place(blue box, table)', 'pick(red box)', 'place(red box, table)', 'pick(hook)', 'push(red box, hook, rack)']\\}}}}}

\noindent\fbox{\parbox{0.97\linewidth}{\small{\texttt{{\\Available scene objects: ['table', 'blue box', 'hook', 'rack', 'red box', 'yellow box']\\
Object relationships: ['on(hook, table)', 'on(red box, table)', 'on(blue box, table)', 'on(yellow box, rack)', 'on(rack, table)']\\
Human instruction: Move the ocean colored box to be under the rack and ensure the hook ends up on the table.\\
Goal predicate set: [['under(blue box, rack)']]\\
Top 1 robot action sequences: ['pick(red box)', 'place(red box, table)', 'pick(yellow box)', 'place(yellow box, rack)', 'pick(hook)', 'push(blue box, hook, rack)', 'place(hook, table)']\\}}}}}

\noindent\fbox{\parbox{0.97\linewidth}{\small{\texttt{{\\Available scene objects: ['table', 'cyan box', 'hook', 'rack', 'red box', 'blue box']\\
Object relationships: ['on(hook, table)', 'on(cyan box, rack)', 'on(rack, table)', 'on(red box, table)', 'inhand(blue box)']\\
Human instruction: How would you set the red box to be the only box on the rack?\\
Goal predicate set: [['on(red box, rack)', 'on(blue box, table)', 'on(cyan box, table)']]\\
Top 1 robot action sequences: ['place(blue box, table)', 'pick(hook)', 'pull(red box, hook)', 'place(hook, table)', 'pick(red box)', 'place(red box, rack)', 'pick(cyan box)', 'place(cyan box, table)']\\}}}}}

\clearpage

\section{Derivations}
\label{appx:derivations}

We provide two derivations to support our approximation of the skill score $S_{\text{skill}}$ (used to select skills while planning with \se{} and \ttm{}) defined in Eq.~\ref{eq:tamp-step-score-factor}.
The skill score is expressed as a product of two terms:
\begin{equation}
    \label{eq:der-tamp-step-score-factor}
    \begin{split}
        S_{\text{skill}}(\psi_t) &= p(\psi_t \mid i, s_1, \psi_{1:t-1}, r_{1:t-1}) \\ 
        &\quad\quad\quad\quad p(r_t \mid i, s_1, \psi_{1:t}, r_{1:t-1}).
    \end{split}
\end{equation}

\subsection{Skill usefulness derivation}
\label{appx-sub:skill-usefulness}
Eq.~\ref{eq:llm-step-score} defines the first term in the skill score product to be the skill \textit{usefulness} score $S_{\text{llm}}$.
We derive the approximation of $S_{\text{llm}}$ given in Eq.~\ref{eq:llm-step-score-decomp}, which corresponds to quantity we use in our experiments.
\begin{align}
    S_{\text{llm}}(\psi_t) 
        &= p(\psi_t \mid i, s_1, \psi_{1:t-1}, r_{1:t-1}) \nonumber \\
        \begin{split}
            &= \int p(\psi_t \mid i, s_{1:t}, \psi_{1:t-1}, r_{1:t-1}) \\
            &\quad\quad\quad\quad p(s_{2:t} \mid i, s_1, \psi_{1:t-1}, r_{1:t-1}) \,d s_{2:t} \nonumber
        \end{split} \\
        &= \E{s_{2:t}}{p(\psi_t \mid i, s_{1:t}, \psi_{1:t-1}, r_{1:t-1})} \label{eq:der-llm-step-score-dep} \\
        &\approx \E{s_{2:t}}{p(\psi_t \mid i, s_{1:t}, \psi_{1:t-1})} \label{eq:der-llm-step-score-indep} \\
        &\approx p(\psi_t \mid i, s_{1:t}, \psi_{1:t-1}) \label{eq:der-llm-step-score-decomp}
\end{align}

The final expression is given in Eq.~\ref{eq:der-llm-step-score-decomp}. Here, we compute a single sample Monte-Carlo estimate of Eq.~\ref{eq:der-llm-step-score-indep} under the future state trajectory $s_{2} \sim T^{\pi_{1}}(\cdot | s_{1}, a^*_{1}), \ldots, s_{t} \sim T^{\pi_{t-1}}(\cdot | s_{t-1}, a^*_{t-1})$, where $a^*_{1:t-1}$ is computed by STAP~\cite{taps-2022}.
The key insight is that future state trajectories $s_{2:t}$ are only ever sampled after STAP has performed geometric feasibility planning to maximize the \textit{success probability} (Eq.~\ref{eq:motion-parameter-score}) of the running plan $\psi_{1:t-1}$.
By doing so, we ensure that the future states $s_{2:t}$ correspond to 
a successful execution of the running plan $\psi_{1:t-1}$, i.e. achieving positive rewards $r_{1:t-1}$.
This supports the independence assumption on rewards $r_{1:t-1}$ used to derive Eq.~\ref{eq:der-llm-step-score-indep} from Eq.~\ref{eq:der-llm-step-score-dep}.

\subsection{Skill feasibility derivation}
\label{appx-sub:skill-feasibility}
Eq.~\ref{eq:motion-step-score} defines the second term in the skill score product (Eq.~\ref{eq:der-tamp-step-score-factor}) as the skill \textit{feasibility} score $S_{\text{geo}}$.
We derive the approximation provided in Eq.~\ref{eq:motion-step-score-decomp}, which is the quantity we use in our experiments.
\begin{align}
    S_{\text{geo}}(\psi_t) 
        &= p(r_t \mid i, s_1, \psi_{1:t}, r_{1:t-1}) \label{eq:der-geo-step-score} \\
        &= p(r_t \mid s_1, \psi_{1:t}, r_{1:t-1}) \label{eq:der-geo-step-score-indep-i} \\
        \begin{split}
            &= \int p(r_t \mid s_{1:t}, \psi_{1:t}, r_{1:t-1}) \\
            &\quad\quad\quad\quad p(s_{2:t} \mid s_1, \psi_{1:t}, r_{1:t-1}) \,d s_{2:t} \nonumber
        \end{split} \\
        &= \E{s_{2:t}}{p(r_t \mid s_{1:t}, \psi_{1:t}, r_{1:t-1})} \label{eq:der-geo-step-score-dep-r} \\
        &\approx \E{s_{2:t}}{p(r_t \mid s_{1:t}, \psi_{1:t})} \label{eq:der-geo-step-score-indep-r} \\
        &= \E{s_{2:t}}{p(r_t \mid s_{1:t}, a^*_{1:t})} \label{eq:der-geo-step-param-score-indep-r} \\
        &= \E{s_{2:t}}{p(r_t \mid s_t, a^*_t)} \label{eq:der-geo-step-param-score-markov} \\
        &= \E{s_{2:t}}{Q^{\pi_t}(s_t, a^*_t)} \label{eq:der-geo-step-Q-score} \\
        &\approx Q^{\pi_t}(s_t, a^*_t) \label{eq:der-geo-step-Q-score-decomp}
\end{align}

From Eq.~\ref{eq:der-geo-step-score} to Eq.~\ref{eq:der-geo-step-score-indep-i}, the reward $r_t$ is conditionally independent of the instruction $i$ given the initial state $s_1$, running plan $\psi_{1:t}$, and previous rewards $r_{1:t-1}$.
As described in Appendix~\ref{appx-sub:skill-usefulness}, we can use STAP to make an independence assumption on the previous rewards $r_{1:t-1}$ between Eq.~\ref{eq:der-geo-step-score-dep-r} and Eq.~\ref{eq:der-geo-step-score-indep-r}. 
The reward probability in Eq.~\ref{eq:der-geo-step-score-indep-r} depends on the parameters $a^*_{1:t}$ computed by STAP and fed to the underlying primitive sequence $\phi_{1:t}$, which gives Eq.~\ref{eq:der-geo-step-param-score-indep-r}.
Eq.~\ref{eq:der-geo-step-param-score-markov} comes from the Markov assumption, and can be reduced to Eq.~\ref{eq:der-geo-step-Q-score} by observing that the reward probability $p(r_t \mid s_t, a^*_t)$ is equal to the Q-value $Q^{\pi_t}(s_t, a^*_t)$ in the contextual bandit setting we consider.
The final expression given in Eq.~\ref{eq:der-geo-step-Q-score-decomp}, which represents a single sample Monte-Carlo estimate of Eq.~\ref{eq:der-geo-step-Q-score} under a sampled future state trajectory $s_{2} \sim T^{\pi_{1}}(\cdot | s_{1}, a^*_{1}), \ldots, s_{t} \sim T^{\pi_{t-1}}(\cdot | s_{t-1}, a^*_{t-1})$.

\color{black}

\clearpage

\section{Real World Demonstration}
\label{appx:demos}

\subsection{Hardware setup}
\label{appx-sub:hardware-setup}
We use a Kinect V2 camera for RGB-D image capture and manually adjust the color thresholds to segment objects in the scene. 
Given the segmentation masks and the depth image, we can estimate object poses to construct the geometric state of the environment. 
\rtwo{For the skill library, we use the same set of policies, Q-functions, and dynamics models trained in simulation.}
We run robot experiments on a Franka Panda robot manipulator.
 
\subsection{Robot demonstration}
Please see our \link{https://sites.google.com/stanford.edu/text2motion}{project page} for demonstrations of \ttm{} operating on a real robot.

\end{appendices}

\end{document}